\documentclass[final,5p,times,twocolumn]{elsarticle}

\usepackage{amssymb}
\usepackage{lipsum}

\usepackage[utf8]{inputenc} 
\usepackage[T1]{fontenc}    
\usepackage{url}            
\usepackage{booktabs}       
\usepackage{amsfonts}       
\usepackage{nicefrac}       
\usepackage{microtype}      
\usepackage{graphicx}       
\usepackage{multirow}       
\usepackage{array}          
\usepackage{lastpage}
\usepackage{changepage}     
\usepackage{xcolor}
\usepackage{hyperref}

\newlength{\extralength}
\newlength{\fulllength}
\definecolor{quantified}{RGB}{0,90,180}    
\definecolor{collection}{RGB}{0,120,90}    
\definecolor{unspecified}{RGB}{120,120,120}  

\journal{Data Science and Management}

\begin{document}

\begin{frontmatter}



\title{Domain Specific Benchmarks for Evaluating Multimodal Large Language Models}


\author[first]{Khizar Anjum\fnref{fn1}}
\author[second]{Muhammad Arbab Arshad\fnref{fn1}}
\author[third]{Kadhim Hayawi}
\author[third]{Efstathios Polyzos}
\author[fourth]{Asadullah Tariq}
\author[fifth]{Mohamed Adel Serhani}
\author[sixth]{Laiba Batool}
\author[seventh]{Brady Lund}
\author[seventh]{Nishith Reddy Mannuru}
\author[seventh]{Ravi Varma Kumar Bevara}
\author[eighth]{Taslim Mahbub}
\author[ninth]{Muhammad Zeeshan Akram}
\author[tenth]{Sakib Shahriar\corref{cor1}}

\fntext[fn1]{These authors contributed equally to this work.}
\cortext[cor1]{Corresponding author.}

\affiliation[first]{organization={Rutgers University},
            city={New Brunswick},
            state={NJ},
            country={USA}}

\affiliation[second]{organization={Iowa State University},
            city={Ames},
            state={IA},
            country={USA}}

\affiliation[third]{organization={Zayed University},
            city={Dubai},
            country={UAE}}

\affiliation[fourth]{organization={United Arab Emirates University},
            city={Al Ain},
            country={UAE}}

\affiliation[fifth]{organization={University of Sharjah},
            city={Sharjah},
            country={UAE}}

\affiliation[sixth]{organization={NUCES},
            city={Karachi},
            country={Pakistan}}

\affiliation[seventh]{organization={University of North Texas},
            city={Denton},
            state={TX},
            country={USA}}

\affiliation[eighth]{organization={George Washington University},
            city={Washington},
            state={DC},
            country={USA}}

\affiliation[ninth]{organization={University of Louisville},
            city={Louisville},
            state={KY},
            country={USA}}

\affiliation[tenth]{organization={University of Guelph},
            city={Guelph},
            state={Ontario},
            country={Canada}}

\begin{abstract}
Large language models (LLMs) are increasingly being deployed across disciplines due to their advanced reasoning and problem-solving capabilities. To measure their effectiveness, various benchmarks have been developed that measure aspects of LLM reasoning, comprehension, and problem-solving. While several surveys address LLM evaluation and benchmarks, a domain-specific analysis remains underexplored in the literature. This paper introduces a taxonomy of seven key disciplines, encompassing various domains and application areas where LLMs are extensively utilized. Additionally, we provide a comprehensive review of LLM benchmarks and survey papers within each domain, highlighting the unique capabilities of LLMs and the challenges faced in their application. Finally, we compile and categorize these benchmarks by domain to create an accessible resource for researchers, aiming to pave the way for advancements toward artificial general intelligence (AGI).
\end{abstract}



\begin{keyword}
Large Language Models \sep Artificial intelligence \sep domain-specific analysis \sep Artificial General Intelligence (AGI) \sep Benchmarking LLMs



\end{keyword}

\end{frontmatter}




\section{Introduction}

The rapid evolution of Multimodal Large Language Models (MLLMs) has led to remarkable general capabilities. However, achieving nuanced performance in specialized areas—the `last mile'—requires a dedicated focus. This paper embarks on a comprehensive survey of domain-specific benchmarks for MLLMs, contextualizing recent advancements, articulating the critical need for domain-specific evaluation, and outlining this survey's main contributions towards fostering targeted MLLM development.

\subsection{Recent Advances and the Rise of Foundational Models}

Recent breakthroughs, exemplified by the launch of highly capable models like GPT-4~\cite{achiam2023gpt} and Gemini~\cite{team2023gemini}, have dramatically shifted the MLLM landscape. Beyond initial scaling in parameters and data, a key advancement lies in ``instruction tuning'' through techniques such as Reinforcement Learning from Human Feedback (RLHF). This has made models more adept at understanding and responding to human requests. Further capabilities include Retrieval Augmented Generation (RAG), enabling MLLMs to incorporate external, up-to-date information, and the development of agentic AI, where models can utilize tools and APIs to perform actions in the digital or physical world. The evolution also extends to more sophisticated reasoning processes, sometimes referred to as ``test-time scaling'' or ``thinking,'' allowing models to tackle complex problems step-by-step. Multimodal capabilities, particularly in areas like image generation and understanding, continue to mature, integrating diverse data types into cohesive reasoning frameworks.

The MLLM revolution is built upon ``foundational models''~\cite{bommasani2021opportunities}, often characterized by their critically central yet incomplete nature and their broad pre-training that allows adaptation to a wide range of downstream tasks. These models, typically based on the transformer architecture (e.g., LLaMA 3~\cite{grattafiori2024llama}, Phi-4~\cite{abdin2024phi}), are initially trained via self-supervised learning on vast and diverse datasets, often encompassing trillions of tokens from text and other modalities. A core principle has been that scaling—increasing model parameters, training data, and computational resources—leads to qualitative shifts and emergent capabilities, such as improved understanding and reasoning, that were not explicitly programmed. These pre-trained foundational models then serve as a base that can be adapted through further training phases, like instruction tuning, to perform a wide array of specific tasks, making them profoundly versatile.

\subsection{The Case for Domain-Specific Benchmarks}

While foundational MLLMs exhibit impressive general intelligence, their efficacy often diminishes when confronted with the nuanced demands of specialized domains. This `last mile problem' underscores a critical gap: general-purpose models, despite their vast training, frequently falter on tasks requiring deep, domain-specific knowledge, intricate reasoning, or precise interpretation of specialized data. For instance, even leading models like GPT-4 have demonstrated significant limitations in specialized financial question answering, achieving low accuracy on benchmarks like FinanceBench~\cite{islam2023financebench}. Similarly, in software engineering, strong performance on general coding tasks does not readily translate to proficiency in domain-specific contexts, as highlighted by evaluations on DomainCodeBench~\cite{zheng2025generalperformancedomain}. Furthermore, in robotics, benchmarks such as MMRo~\cite{li2024mmro} reveal that while models may excel at high-level planning, they can struggle with fundamental perceptual tasks crucial for real-world interaction. These examples, which are explored in detail in subsequent sections of this paper, motivate the central thesis of this survey: the critical need for developing and analyzing domain-specific benchmarks to truly harness and evaluate the potential of MLLMs in diverse, real-world applications.

This focus on domain specificity is not limited to academic research; it is also evident in the trajectory of leading commercial MLLMs. For instance, models like Anthropic's Claude have demonstrated strong performance in software engineering tasks, as seen on benchmarks such as SWE-bench Verified where Claude 3 Opus achieved 70.3\%, slightly ahead of OpenAI's GPT-4o (referred to as `o3') at 69.1\%. Conversely, on mathematical reasoning benchmarks like AIME 2025 (a proxy for high-level math contests), GPT-4o showed a significant lead with 88.9\% compared to Claude 3 Opus's 49.5\%. This divergence highlights that even large foundational models exhibit varying strengths and weaknesses across different domains, reinforcing the idea that a one-size-fits-all approach has limitations and that specialized capabilities are becoming key differentiators.

Given these limitations of generalist models, this survey turns its attention to the growing ecosystem of benchmarks specifically engineered for domain-focused MLLM development. These specialized benchmarks are not merely for assessing performance within a niche; they are crucial for driving innovation and refining models to handle the unique challenges of individual fields. For example, benchmarks in robotics like MMRo~\cite{li2024mmro} serve as diagnostic tools, pinpointing failures in perception that general MLLMs must overcome for effective real-world deployment. Similarly, evaluations in areas like software engineering with DomainCodeBench~\cite{zheng2025generalperformancedomain} highlight the necessity for domain-aware prompting and contextual understanding, techniques which can enhance MLLM applicability more broadly. By pushing MLLMs to master complex, domain-specific data and reasoning patterns—from understanding financial nuances to interpreting specialized engineering diagrams or medical imagery—these benchmarks also provide invaluable feedback. The insights and architectural advancements spurred by the need to succeed on these focused evaluations can, in turn, enrich the development of more robust and versatile foundational models, contributing to overall MLLM progress.

\begin{figure*}[t]
  \centering
  \includegraphics[width=\fulllength]{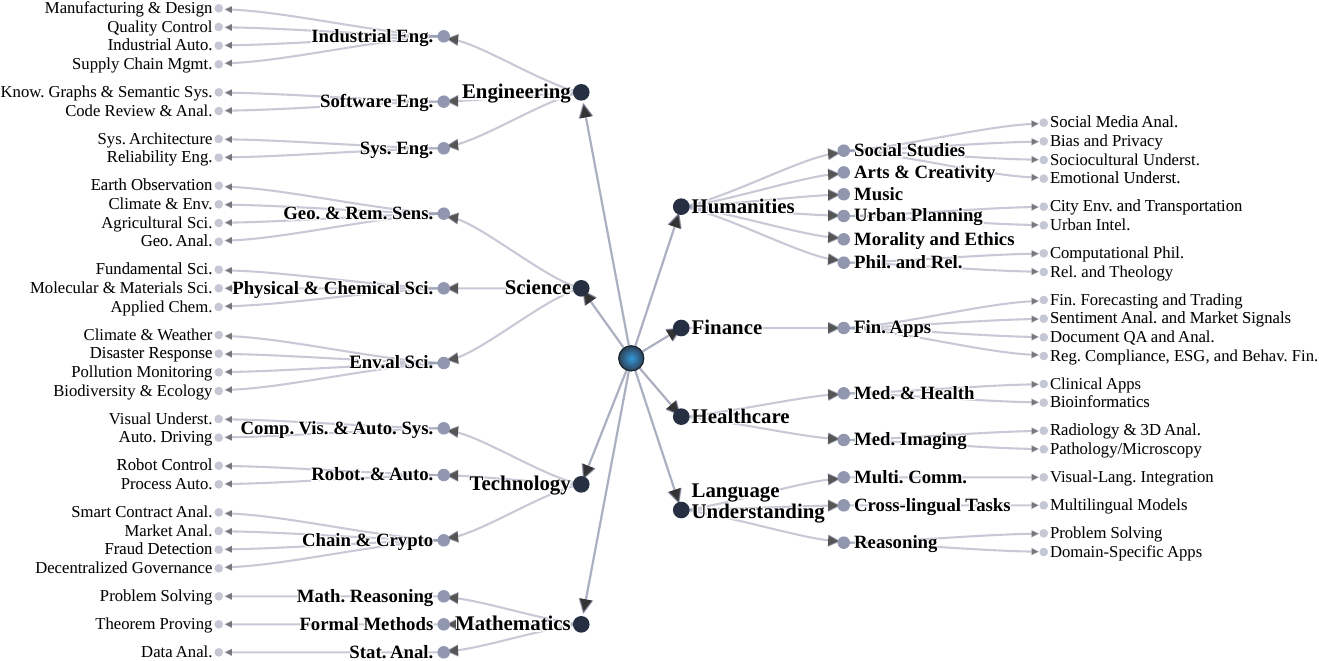}
  \caption{The domain hierarchy of the eight disciplines covered in this article. The hierarchy is based on the domains and sub-domains of each discipline, and the sub-domains are further broken down into their application areas.}
  \label{fig:domain_hierarchy}
\end{figure*}

\subsubsection{Main Contributions}
Our main contributions are as follows:
\begin{itemize}
    \item Presents a pioneering and comprehensive survey of domain-specific benchmarks for Multimodal Large Language Models (MLLMs) across a wide array of disciplines.
    \item Highlights the critical role of developing MLLMs tailored for diverse domains as essential for addressing the `last mile problem' and achieving practical efficacy in real-world applications.
    \item Argues that such specialized advancements not only ensure domain-specific efficacy but also supplementally contribute to the broader development and refinement of large foundational models.
\end{itemize}

\section{Review Methodology \& Framework}

This paper presents a systematic review of domain-specific benchmarks for Multimodal Large Language Models (MLLMs). Our review examines \textbf{eight key disciplines}, which are explored in detail in Sections \ref{sec:engineering}-\ref{sec:language}:

\begin{enumerate}
    \item \textbf{Engineering} (Section \ref{sec:engineering}) - Including industrial engineering, software engineering, and systems engineering
    \item \textbf{Science} (Section \ref{sec:science}) - Covering geography \& remote sensing, physics \& chemistry, and environmental science
    \item \textbf{Technology} (Section \ref{sec:technology}) - Encompassing computer vision \& autonomous systems, robotics \& automation, and blockchain \& cryptocurrency
    \item \textbf{Mathematics} (Section \ref{sec:mathematics}) - Addressing mathematical reasoning, formal methods, and statistical analysis
    \item \textbf{Humanities} (Section \ref{sec:humanities}) - Examining social studies, arts \& creativity, music, urban planning, morality \& ethics, and philosophy \& religion
    \item \textbf{Finance} (Section \ref{sec:finance}) - Covering financial forecasting, sentiment analysis, document QA, and regulatory compliance
    \item \textbf{Healthcare} (Section \ref{sec:healthcare}) - Focusing on medicine \& healthcare and medical imaging
    \item \textbf{Language Understanding} (Section \ref{sec:language}) - Including multimodal communication, cross-lingual tasks, and reasoning
\end{enumerate}

Each discipline is further broken down into specific domains, their respective sub-domains and application areas, as visually outlined in Figure~\ref{fig:domain_hierarchy}. A core component of our analysis for each domain is a comprehensive table designed to consolidate critical information about relevant benchmarks and survey papers. These tables meticulously detail characteristics such as:

\begin{itemize}
    \item \textbf{Scale}: Dataset size and diversity, color-coded to indicate \textcolor{quantified}{precise quantities} (blue), \textcolor{collection}{collection descriptors} (green), or \textcolor{unspecified}{unspecified amounts} (gray)
    \item \textbf{Task Type}: Nature and complexity of the evaluation task
    \item \textbf{Input Modality}: Input types utilized (text, image, audio, video)
    \item \textbf{Model}: Models evaluated in the benchmark
    \item \textbf{Performance}: Quantitative or qualitative results reported
    \item \textbf{Key Focus}: Core objectives and applications of the benchmark
\end{itemize}

Complementing the structured information in the tables, the textual discussion within each domain section offers deeper qualitative insights. This narrative focuses on elucidating overarching trends, significant advancements, and persistent challenges observed within the various application areas. By synthesizing these observations, we aim to provide a nuanced understanding that extends beyond the summarized data, highlighting the current capabilities and limitations of MLLMs in specialized contexts and explaining the significance of each referenced paper in advancing the field.

\textbf{Search Strategy and Scope.}
Our search strategy involved systematically querying major academic databases (e.g., IEEE Xplore, ACM Digital Library, Scopus) and preprint archives (e.g., arXiv) using a combination of general keywords such as ``Large Language Model,'' ``Multimodal LLM,'' ``benchmark,'' ``evaluation,'' and domain-specific terms relevant to the disciplines covered. The scope of this review encompasses studies focused on Large Language Models (LLMs), Vision Language Models (VLMs), and more broadly, Multimodal Large Language Models (MLLMs). We specifically prioritized papers that introduce new benchmarks, conduct empirical evaluations of MLLMs on existing benchmarks, or provide comprehensive analyses of model performance within particular domains. Consequently, purely theoretical papers without a significant evaluation component were generally excluded. Survey papers are selectively included within specific sub-domains if they offer a valuable consolidation of existing benchmarks or evaluation practices pertinent to that application area, thereby enriching the contextual understanding of the domain's evaluation landscape. Our primary focus is on papers that contribute to understanding how MLLMs are evaluated and perform in various specialized fields, rather than theoretical explorations of model architectures or training methodologies alone.

\section{Engineering}\label{sec:engineering}
Large Language Models (LLMs) have emerged as transformative tools across engineering disciplines, revolutionizing how engineers approach complex problems, design systems, and manage technical knowledge. By leveraging their ability to understand and generate domain-specific content, LLMs are increasingly being integrated into engineering workflows to enhance productivity, facilitate knowledge transfer, and support decision-making processes. The following sections explore how various engineering disciplines are developing specialized benchmarks to evaluate and improve LLM performance in technical contexts, highlighting both the promising capabilities and current limitations of these models in addressing real-world engineering challenges. An overview of all the relevant benchmarks is presented in Table~\ref{tbl:engineering_discipline}.

\begin{table*}[!ht]
  \caption{Engineering Discipline: Domain-Specific Benchmarks}\label{tbl:engineering_discipline}
  \centering
  
  \resizebox{\fulllength}{!}{%
  \begin{tabular}{lllcccccc}
  \cmidrule[\heavyrulewidth]{1-9}
  \textbf{Domain} & \textbf{Sub-domain} & \textbf{Benchmark} & \textbf{Scale} & \textbf{Task Type} & \textbf{Input Modality} & \textbf{Model} & \textbf{Performance} & \textbf{Key Focus} \\
  \cmidrule[\heavyrulewidth]{1-9}
  \multirow{12}{*}{Industrial Engineering} 
  & \multirow{4}{*}{Manufacturing \& Design}
  & DesignQA~\cite{designqa} & \textcolor{unspecified}{N/A} & Rule Comprehension, Extraction & Text + CAD Images & Multiple & 1,449 Qs & Engineering Documentation \\
  && Freire et al.~\cite{kernan2024knowledge} & \textcolor{collection}{Multiple tasks} & Knowledge Retrieval & Text & Multiple & 97.5\% factuality & Factory Documentation \\
  && Manu-Eval~\cite{liu2024manu} & \textcolor{quantified}{22 subcategories} & Manufacturing Tasks & Multi-modal & 20 LLMs & \textcolor{unspecified}{N/A} & Manufacturing Industry \\
  && FDM-Bench~\cite{eslaminia2024fdm} & \textcolor{collection}{Multiple tasks} & FDM-specific Tasks & Text + G-code & Multiple & 62\% accuracy & Additive Manufacturing \\
  \cmidrule{2-9}
  & \multirow{1}{*}{Quality Control}
  & LLM4PLC~\cite{10.1145/3639477.3639743} & \textcolor{collection}{Multiple tasks} & Code Generation & Text & Multiple & 72\% pass rate & PLC Programming \\
  \cmidrule{2-9}
  & \multirow{3}{*}{Industrial Automation}
  & Tizaoui et al.~\cite{tizaoui2024towards} & \textcolor{collection}{Multiple tasks} & Extractive QA & Text & 6 LLMs & \textcolor{unspecified}{N/A} & Process Automation \\
  && Xia et al.~\cite{xia2024incorporating} & \textcolor{collection}{Multiple tasks} & Production Planning & Multi-modal & LLM Agents & \textcolor{unspecified}{N/A} & Production Systems \\
  && Ogundare et al.~\cite{ogundare2023industrial} & \textcolor{collection}{Multiple tasks} & Problem Solving & Text & ChatGPT & \textcolor{unspecified}{N/A} & Oil \& Gas Engineering \\
  \cmidrule{2-9}
  & \multirow{3}{*}{Supply Chain Management}
  & Rahman et al.~\cite{rahman2024leveraging} & \textcolor{collection}{Multiple tasks} & Optimization & Text & GPT-4o & 95\% correct & Supply Chain Optimization \\
  && OptiGuide~\cite{li2023scm} & \textcolor{collection}{Multiple tasks} & Query Processing & Text & LLMs & 93\% accuracy & Supply Chain Operations \\
  && Raman et al.~\cite{raman2024ai} & \textcolor{quantified}{150 questions} & Question Answering & Text & ChatGPT, Bard & 4.95/5 accuracy & Supply Chain Management \\
  \cmidrule[\heavyrulewidth]{1-9}
  \multirow{6}{*}{Software Engineering} 
  & \multirow{2}{*}{\begin{tabular}[c]{@{}l@{}}Knowledge Graphs \\ \& Semantic Systems\end{tabular}}
  & LLM-KG-Bench~\cite{llm-kg-bench} & \textcolor{collection}{Multiple tasks} & Knowledge Graph Engineering & Text & Multiple & \textcolor{unspecified}{N/A} & Knowledge Graph Generation \\
  && BIG-bench~\cite{srivastava2023beyond} & \textcolor{quantified}{204 tasks} & Multiple Tasks & Text & Multiple & \textcolor{unspecified}{N/A} & General Capabilities \\
  \cmidrule{2-9}
  & \multirow{4}{*}{Code Review \& Analysis}
  & Azanza et al.~\cite{azanza2025trackingmovingtargetframework} & \textcolor{collection}{Multiple tasks} & Test Generation & Text & Multiple & 90\% (2024) & Software Testing \\
  && DomainCodeBench~\cite{zheng2025generalperformancedomain} & \textcolor{quantified}{2,400 tasks} & Code Generation & Text & 10 LLMs & +38\% w/ context & Domain-specific Coding \\
  && StackEval~\cite{shah2024stackeval} & \textcolor{collection}{Multiple tasks} & Coding Assistance & Text & Multiple & 95.5\% (hist.), 83\% (unseen) & Code Writing \& Review \\
  && Zhang et al.~\cite{zhang2024surveylargelanguagemodels} & \textcolor{quantified}{947 studies} & Multiple Tasks & Text & 62 LLMs & \textcolor{unspecified}{N/A} & Software Engineering \\
  \cmidrule[\heavyrulewidth]{1-9}
  \multirow{4}{*}{Systems Engineering} 
  & \multirow{1}{*}{System Architecture}
  & SysEngBench~\cite{bell2024introducing} & \textcolor{collection}{Multiple tasks} & Systems Engineering Tasks & Text & Multiple & \textcolor{unspecified}{N/A} & Systems Engineering \\
  \cmidrule{2-9}
  & \multirow{3}{*}{Reliability Engineering}
  & Hu et al.~\cite{hu2024use} & \textcolor{collection}{Multiple tasks} & Question Answering & Text & GPT-4, GPT-3.5 & 92\% (CRE) & Reliability Engineering \\
  && Platinum Benchmarks~\cite{vendrow2025large} & \textcolor{quantified}{15 benchmarks} & Multiple Tasks & Text & Multiple & \textcolor{unspecified}{N/A} & Model Reliability \\
  && Liu et al.~\cite{liu2024trustworthyllmssurveyguideline} & \textcolor{quantified}{29 categories} & Trustworthiness Assessment & Text & Multiple & \textcolor{unspecified}{N/A} & LLM Trustworthiness \\
  \cmidrule[\heavyrulewidth]{1-9}
  \end{tabular}%
  }

\end{table*}

\subsection{Industrial Engineering}
In the realm of industrial engineering, LLMs are being deployed to optimize manufacturing processes, improve quality control, streamline supply chain operations, and enhance industrial automation. These applications require models to interpret technical specifications, understand engineering drawings, analyze production data, and generate actionable insights. Industrial engineering benchmarks are particularly focused on evaluating how well LLMs can bridge the gap between general language understanding and specialized domain knowledge, with emphasis on their ability to reason about physical systems, interpret multimodal inputs (text, images, CAD files), and provide reliable recommendations in safety-critical environments. The growing adoption of LLMs in this field reflects their potential to address persistent challenges in knowledge management, process optimization, and decision support across various industrial sectors.

\subsubsection{Manufacturing \& Design}
Recent years have seen the emergence of several specialized benchmarks and studies aimed at rigorously evaluating the capabilities of large language models (LLMs) and multimodal LLMs (MLLMs) in the context of manufacturing and design~\cite{designqa,eslaminia2024fdm,kernan2024knowledge}. The following paragraphs provide detailed explanations of key works in this area, each highlighting unique contributions, evaluation strategies, and the current state of LLM performance in addressing real-world engineering challenges.

DesignQA~\cite{designqa} introduces a novel multimodal benchmark for evaluating the ability of large language models (LLMs) and multimodal LLMs (MLLMs) to comprehend and apply engineering requirements from technical documentation. Developed using real-world data from the Formula SAE competition and the MIT Motorsports team, DesignQA uniquely combines textual design requirements, CAD images, and engineering drawings. The benchmark is structured into three segments—Rule Extraction, Rule Comprehension, and Rule Compliance—mirroring the core tasks engineers face in design processes. It features 1,449 questions that test models on extracting rules from lengthy documents, identifying components in CAD models, and checking design compliance with requirements. Evaluation of state-of-the-art models (GPT-4o, GPT-4, Claude-Opus, Gemini-1.0, LLaVA-1.5) revealed that, while GPT-4o generally performed best, all models struggled with reliably extracting rules, recognizing technical components, and analyzing engineering drawings. The study highlights the current limitations of MLLMs in handling complex engineering documentation and sets a foundation for future research in AI-assisted engineering design.

FDM-Bench~\cite{eslaminia2024fdm} is the first comprehensive benchmark specifically designed to evaluate LLMs on tasks related to Fused Deposition Modeling (FDM) in additive manufacturing. Developed by a collaboration of researchers from the University of Illinois, Rutgers, and the University of Michigan, FDM-Bench assesses both the ability of LLMs to detect anomalies in G-code (the programming language for 3D printers) and to answer user queries at varying expertise levels. The benchmark includes both multiple-choice and open-ended questions, as well as G-code samples representing different types of print defects. Four leading LLMs—GPT-4o, Claude 3.5 Sonnet, Llama-3.1-70B, and Llama-3.1-405B—were evaluated. Results showed that closed-source models (GPT-4o, Claude) generally outperformed open-source models in G-code anomaly detection, while Llama-3.1-405B performed comparably or better in user query response. The study demonstrates the promise of LLMs in supporting FDM 3D printing but also highlights the need for further improvements, especially in visual defect detection and domain-specific fine-tuning.

Freire et al.~\cite{kernan2024knowledge} present a practical LLM-based knowledge management system for manufacturing environments, focusing on improving information retrieval and knowledge sharing among factory operators. The system uses Retrieval Augmented Generation (RAG) to answer operator queries by pulling relevant information from factory manuals and issue reports. A user study conducted at a detergent factory revealed that the tool was easy to use and improved access to information, but users still preferred learning from human experts when possible. The study also benchmarked seven LLMs (including GPT-4, GPT-3.5, Mixtral 8x7B, Llama 2, StableBeluga2, and Guanaco variants) on their ability to answer factory-specific questions. GPT-4 achieved the highest factuality and completeness, but newer open-source models like StableBeluga2 and Mixtral 8x7B showed strong performance and lower hallucination rates, making them attractive for privacy-sensitive industrial applications. The research highlights both the potential and the challenges of deploying LLM-powered tools in manufacturing, especially regarding safety, user acceptance, and the need for human oversight.

\subsubsection{Quality Control}
Quality control in industrial settings presents unique challenges for LLMs, particularly in the context of code generation and validation for programmable logic controllers (PLCs). LLM4PLC~\cite{10.1145/3639477.3639743} presents an innovative pipeline for leveraging Large Language Models (LLMs) to generate verifiable programming code for Programmable Logic Controllers (PLCs) in industrial control systems. The approach integrates external verification tools—including grammar checkers, compilers, and symbolic model verifiers—into a user-guided, iterative workflow that transforms natural language specifications into IEC 61131-3 Structured Text code. The pipeline features automated feedback loops, prompt engineering, and LoRA-based fine-tuning to iteratively improve code quality and correctness. Experimental results on a real manufacturing testbed show that the pipeline increases code pass rates from 47\% to 72\% and boosts expert-rated code quality, while also reducing engineering effort from hours to minutes. The study demonstrates that, with proper verification and adaptation, LLMs can bridge the gap between natural language requirements and reliable, deployable PLC code, though challenges remain in ensuring explainability and safe deployment in critical systems.

\subsubsection{Industrial Automation}
Recent research in industrial automation has produced a range of benchmarks and case studies that explore the integration of large language models (LLMs) into process automation, production planning, and complex engineering problem-solving. The following paragraphs summarize key contributions in this area, highlighting the development of domain-specific datasets, the use of LLM agents in digital twin systems, and the evaluation of LLMs in specialized industrial contexts.

Tizaoui et al.~\cite{tizaoui2024towards} address the lack of domain-specific datasets for process automation by introducing a novel extractive question answering (QA) benchmark. Their methodology combines automated and manual techniques to generate and annotate a diverse set of QA pairs from academic papers in process automation. The resulting dataset, which includes both original and paraphrased questions, is used to fine-tune and evaluate six encoder-only LLMs. The study demonstrates that fine-tuning and data augmentation significantly improve model performance, and that models generalize well to unseen documents within the domain. This work establishes a foundation for future industrial NLP efforts by providing a robust benchmark and evaluation framework for process automation tasks.

Xia et al.~\cite{xia2024incorporating} propose a novel approach to integrating LLMs into automated production systems using a hierarchical framework based on the automation pyramid. Their system semantically enriches low-level production data, making it interpretable for LLM agents that generate process plans and decompose them into atomic operations executed as microservices. The authors implement a digital twin system and a multi-agent LLM architecture in a modular production facility, demonstrating the feasibility of LLM-driven planning and control. While the prototype shows promise in generating executable commands and handling both predefined and unforeseen scenarios, the study notes challenges related to real-time performance, the need for comprehensive real-world testing, and the importance of cost-benefit analysis for industrial adoption.

Ogundare et al.~\cite{ogundare2023industrial} evaluate the capabilities and limitations of LLMs, particularly ChatGPT, in solving complex oil and gas engineering problems. The paper assesses LLM performance on mathematical modeling tasks such as fluid flow, reservoir simulation, and equation discretization, using both stepwise and aggregate quality metrics. While LLMs demonstrate strong theoretical reasoning and the ability to generate sound equations for standard problems, they struggle with non-standard scenarios, complex geometries, and computational accuracy. The authors recommend enhancing LLMs with domain-specific knowledge and physical constraints, and suggest that LLMs are most valuable for theoretical model development rather than direct numerical computation in industrial engineering.

\subsubsection{Supply Chain Management}
Recent advances in supply chain management (SCM) and operations research have leveraged large language models (LLMs) to automate optimization, enhance interpretability, and support both educational and professional decision-making. The following paragraphs summarize key contributions in this area, highlighting frameworks for code generation and solver integration, natural language interfaces for optimization, and comparative evaluations of generative AI tools in SCM contexts.

Rahman et al.~\cite{rahman2024leveraging} demonstrate how LLMs, specifically GPT-4o, can bridge the gap between natural language problem descriptions and mathematical optimization in supply chain management. Their two-stage framework translates SCM problems into executable GUROBI code and interprets solver outputs for user-friendly analysis, using prompt engineering and advanced reasoning techniques like Tree of Thoughts. The study shows that LLMs can generate error-free code and provide accurate interpretations for a range of transportation problems, achieving approximately 95\% correctness in output interpretation. While the approach accelerates model development and enhances accessibility, the authors note limitations in handling highly complex problems and emphasize the need for expert verification and further industrial validation.

Li et al.~\cite{li2023scm} introduce OptiGuide, a privacy-preserving framework that integrates LLMs with combinatorial optimization solvers to provide natural language explanations of supply chain decisions. OptiGuide enables users to pose plain text queries, which are translated into optimization code and executed by solvers, with results interpreted back into human-readable explanations. The system is designed to work alongside existing optimization technology, preserving proprietary data and leveraging in-context learning for flexible adaptation. Deployed in Microsoft's cloud supply chain, OptiGuide achieved over 90\% accuracy in real-world server placement scenarios and received positive feedback from planners and engineers. The authors highlight the importance of specificity in user queries and the need for robust application-specific safeguards.

Raman et al.~\cite{raman2024ai} provide a comparative analysis of ChatGPT and Bard in the context of supply chain management education, using a dataset of 150 certified supply chain professional exam questions. The study evaluates both models on accuracy, relevance, clarity, and readability, finding that ChatGPT outperforms Bard in accuracy and relevance, while Bard demonstrates slightly better readability. Both tools are shown to be effective study aids and professional resources, capable of supporting lesson planning, practice question development, and corporate training. The authors suggest that continual model adaptation and the use of diverse datasets are essential for ensuring ethical, effective, and up-to-date AI applications in supply chain management.

\subsection{Software Engineering}
Software Engineering has emerged as one of the most promising application domains for LLMs, with models demonstrating remarkable capabilities in code generation, debugging, documentation, and knowledge management. Benchmarks in this field evaluate LLMs on their ability to understand programming languages, follow software specifications, adhere to best practices, and integrate with existing codebases. These evaluations span diverse programming paradigms, languages, and development environments, reflecting the multifaceted nature of modern software engineering. While LLMs show impressive performance in generating syntactically correct code and explaining programming concepts, challenges remain in ensuring semantic correctness, security, and maintainability of the generated solutions. The following sections explore specialized benchmarks that assess LLM performance in knowledge representation and code analysis tasks that are fundamental to software engineering practice.

\subsubsection{Knowledge Graphs \& Semantic Systems}
Recent work in knowledge graph engineering and semantic systems has led to the development of specialized benchmarks for evaluating large language models (LLMs) on tasks such as knowledge extraction, syntax correction, and semantic reasoning. The following paragraphs summarize key contributions in this area, highlighting frameworks for automated evaluation and the breadth of tasks used to probe LLM capabilities.

LLM-KG-Bench~\cite{llm-kg-bench} introduces a modular benchmarking framework specifically designed to assess LLMs in knowledge graph engineering (KGE) tasks. The framework features automated evaluation modules for tasks such as fixing errors in Turtle files, extracting knowledge from plaintext, and generating synthetic datasets. Empirical results show that while LLMs like GPT-4 and Claude-1.3 outperform earlier models in syntax and extraction tasks, all models struggle with zero-shot knowledge graph generation and show diminishing performance as task complexity increases. LLM-KG-Bench provides a scalable, automated approach for tracking LLM performance in KGE and highlights both the potential and current limitations of LLMs in semantic web applications.

BIG-bench~\cite{srivastava2023beyond} is a large-scale, collaborative benchmark designed to evaluate and extrapolate the capabilities of LLMs across 204 diverse tasks, including linguistics, reasoning, software development, and more. The benchmark was created to address the rapid obsolescence of existing benchmarks as models improve, and it includes contributions from hundreds of researchers worldwide. Key findings indicate that while model performance and calibration improve with scale, even the largest models lag behind human experts on many tasks, and performance can be brittle and highly sensitive to prompt formulation. BIG-bench remains a living benchmark, continually updated to track the evolving capabilities and limitations of LLMs in a wide range of semantic and reasoning tasks.

\subsubsection{Code Review \& Analysis}
Recent research in code review and analysis has produced a variety of benchmarks and frameworks to evaluate large language models (LLMs) on tasks such as code generation, debugging, test generation, and software engineering lifecycle support. The following paragraphs summarize key contributions in this area, highlighting the importance of domain-specific evaluation, continuous benchmarking, and the integration of LLMs into real-world software engineering workflows.

DomainCodeBench~\cite{zheng2025generalperformancedomain} introduces a multi-domain benchmark designed to assess LLM performance on code generation tasks across 12 application domains and 15 programming languages. The study reveals that strong performance on general coding benchmarks does not guarantee success in specialized domains, as LLMs often struggle with domain-specific libraries, algorithms, and workflows. Contextual enhancement—such as providing relevant code snippets—significantly improves results, underscoring the need for domain-aware evaluation and prompt design in practical development scenarios.

StackEval~\cite{shah2024stackeval} presents a comprehensive benchmark for coding assistance, featuring two datasets: StackEval, which covers 925 curated Stack Overflow questions across 25 languages and four task types, and StackUnseen, which evaluates LLMs on emerging technologies and recent content. The benchmark also introduces an LLM-as-a-Judge framework to assess the quality of generated code. Results show that while top models excel on historical data, performance drops on new or unseen problems, highlighting the need for continuous adaptation and responsible AI development in coding assistance.

Azanza et al.~\cite{azanza2025trackingmovingtargetframework} propose a measurement framework for the continuous evaluation of LLM-based test generators in industrial environments. Validated through a longitudinal study at a software consultancy, the framework integrates with industry tools like SonarQube and tracks metrics such as code quality, coverage, and test parameterization. The study demonstrates rapid improvements in LLM-generated test quality over time, but emphasizes the ongoing need for expert oversight, prompt engineering, and systematic re-evaluation as models evolve.

Zhang et al.~\cite{zhang2024surveylargelanguagemodels} provide a comprehensive survey of LLM applications in software engineering, categorizing 947 studies across 112 SE tasks and five phases of the software lifecycle. The survey highlights the transformative impact of LLMs on requirements, development, testing, maintenance, and management, while also discussing challenges related to evaluation, security, domain adaptation, and model optimization. The authors advocate for the creation of clean, diverse benchmarks and the integration of explainable AI techniques to advance LLM-based software engineering research and practice.

\subsection{Systems Engineering}
Systems engineering integrates multiple disciplines to design and manage complex systems throughout their lifecycles. Large language models (LLMs) have emerged as valuable tools in this domain, supporting critical tasks from requirements elicitation to stakeholder communication. Their natural language capabilities help bridge communication gaps between technical and non-technical stakeholders—a persistent challenge in systems engineering. Recent benchmarks like SysEngBench~\cite{bell2024introducing} and platinum benchmarks.~\cite{vendrow2025large} evaluate LLMs' effectiveness in model-based systems engineering and safety-critical verification. As shown in Table~\ref{tbl:engineering_discipline}, LLMs accelerate documentation, enhance knowledge integration, and improve requirements management, though challenges remain in handling interdisciplinary complexity and ensuring reliability in mission-critical applications.

\subsubsection{Systems Architecture}
In the context of systems architecture, LLMs are evaluated on their capacity to support the design and orchestration of complex, hierarchical production systems. For example, SysEngBench~\cite{bell2024introducing} introduces a novel benchmark for assessing LLMs in systems engineering, with a focus on model-based systems engineering (MBSE) and the transition from document-centric to model-centric approaches. The benchmark covers ten core topic areas using multiple-choice questions derived from lecture slides and human review, and evaluates models such as Llama 2, Mistral, and Orca 2. Results show that while some models perform well in certain areas, challenges remain in requirements analysis and in handling complex, real-world systems engineering tasks. SysEngBench provides a foundation for future research, with plans to expand the benchmark to include more complex questions, multimodal inputs, and human performance comparisons.

\subsubsection{Reliability Engineering}
Recent work in reliability engineering has produced a range of studies and benchmarks to evaluate the trustworthiness, reliability, and alignment of large language models (LLMs) in high-stakes engineering contexts. The following paragraphs summarize key contributions in this area, highlighting empirical evaluations, the development of platinum benchmarks, and comprehensive trustworthiness frameworks.

Hu et al.~\cite{hu2024use} examine the application of LLMs in reliability engineering by testing GPT-3.5 and GPT-4 on Certified Reliability Engineer (CRE) exam questions. The study finds that prompt engineering and context-specific instructions significantly improve model performance, with GPT-4 achieving up to 92\% accuracy. While LLMs show promise for information extraction, FMEA processing, and code generation, the authors caution against sole reliance on LLMs for safety-critical or regulatory decisions, emphasizing the need for human oversight and collaboration.

Platinum Benchmarks~\cite{vendrow2025large} addresses the gap between capability and reliability in current LLM benchmarks by proposing ``platinum benchmarks''—carefully curated to minimize label errors and ambiguity. The study demonstrates that even frontier models make simple mistakes on basic tasks, and that existing benchmarks often fail to measure true reliability due to pervasive label errors. The authors advocate for rigorous reliability testing as a standard practice for deployed LLMs, similar to traditional reliability engineering.

Liu et al.~\cite{liu2024trustworthyllmssurveyguideline} present a comprehensive framework for evaluating LLM trustworthiness and alignment, introducing a taxonomy of seven major categories and 29 sub-categories. Their empirical studies reveal that while more aligned models generally perform better, alignment effectiveness varies across trustworthiness dimensions, and even well-aligned models have limitations in fairness, causal reasoning, and robustness. The authors highlight the need for fine-grained, targeted alignment strategies and automated evaluation methods to ensure LLMs meet societal values and ethical standards.

\section{Science}\label{sec:science}

The scientific domain presents unique challenges for evaluating large language models, requiring deep domain expertise, multi-step reasoning, and the ability to work with diverse data modalities. This section explores how LLMs are being evaluated and applied across geography, physical sciences, and other scientific disciplines. An overview of all the relevant benchmarks is presented in Table~\ref{tbl:science_discipline}.

\begin{table*}[!ht]
  \caption{Science Discipline: Domain-Specific Benchmarks}\label{tbl:science_discipline}
  \centering
  
  \resizebox{\fulllength}{!}{%
  \begin{tabular}{lllcccccc} 
  \cmidrule[\heavyrulewidth]{1-9}
  \textbf{Domain} & \textbf{Sub-domain} & \textbf{Benchmark} & \textbf{Scale} & \textbf{Task Type} & \textbf{Input Modality} & \textbf{Model} & \textbf{Performance} & \textbf{Key Focus} \\
  \cmidrule[\heavyrulewidth]{1-9}
  Geography & \multirow{4}{*}{Earth Observation}
  & TEOChat~\cite{irvin_teochat_2024} & \textcolor{unspecified}{N/A} & Temporal EO & Image sequence & \textit{TEOChat} & SOTA & Earth observation \\
  \& Remote && EarthNets~\cite{xiong_earthnets_2024} & \textcolor{collection}{500+ datasets} & Earth obs & Multi-modal & Multiple & \textcolor{unspecified}{N/A} & Dataset benchmark \\
  Sensing && VLEO-Bench~\cite{zhang_good_2024} & \textcolor{unspecified}{N/A} & EO analysis & Satellite & GPT-4V & Mixed & Scene understanding \\
  \cmidrule{2-9}
  & \multirow{1}{*}{Climate \& Environment}
  & STBench~\cite{li_stbench_2024} & \textcolor{quantified}{60K QA pairs} & Spatio-temporal & Multi-modal & Multiple & \textcolor{unspecified}{N/A} & Data mining \\
  \cmidrule{2-9}
  & \multirow{1}{*}{Agricultural Science}
  & AgriLLM~\cite{sapkota_multi-modal_2024} & Survey paper & Agriculture & Multi-modal & Multiple & \textcolor{unspecified}{N/A} & Farming applications \\
  \cmidrule{2-9}
  & \multirow{4}{*}{Geospatial Analysis}
  & GEOBench-VLM~\cite{danish_geobench-vlm_2024} & \textcolor{quantified}{10K instructions} & Geospatial & Multi-modal & GPT-4V & 40\% & Remote sensing \\
  && Roberts et al.~\cite{roberts_charting_2024} & \textcolor{unspecified}{N/A} & Geographic & Multi-modal & GPT-4V & Human-level & Geospatial tasks \\
  && RSUniVLM~\cite{liu_rsunivlm_2024} & \textcolor{quantified}{1B params} & Multi-granular & Multi-modal & \textit{RSUniVLM} & SOTA & Remote sensing \\
  && INS-MMBench~\cite{lin_ins-mmbench_2024} & \textcolor{quantified}{2.2K questions} & Insurance & Multi-modal & Multiple & \textcolor{unspecified}{N/A} & Domain tasks \\
  \cmidrule[\heavyrulewidth]{1-9}
  Physical & \multirow{10}{*}{Fundamental Science} 
  & MMLU \cite{hendrycks2020measuring} & \textcolor{collection}{Varies by Domain} & QA & Text & Multiple & Below Expert-Level & Academic knowledge \\
  \& Chemical && ScienceQA \cite{lu2022learn} & \textcolor{collection}{Varies by Domain} & QA & Text and Images & GPT3 with CoT & 75.17\% & High-School QA \\
  Sciences && IsoBench \cite{fu2024isobench} & \textcolor{quantified}{75 per domain} & Multimodal QA & Text and Images & Multiple & Text better than Image & General Science \\
  && VisScience \cite{jiang2024visscience} & \textcolor{quantified}{1000 per domain} & Multi-modal reasoning & Text and Images & Multiple & 38.2\% Physics, 47.0\% Chemistry & K12 assessment \\
  && GPQA \cite{rein2024gpqa} & \textcolor{quantified}{448 MCQs} & Difficult MCQ & Text & Claude 3 Opus & PhD-level (60\%) & Graduate-level reasoning \\
  && MM-PhyQA \cite{anand2024mm} & \textcolor{quantified}{4500 q/a} & Q/A & Text and Images & LLaVA-1.5 13b & 71.65\% & High-school physics \\
  && ChemBench \cite{mirza2024large} & \textcolor{quantified}{2700+ q/a pairs} & Q/A, Reasoning & Text & Multiple & 64\% accuracy & General Chemistry \\
  && ChemQA \cite{chemQA2024} & \textcolor{quantified}{85k examples} & Chemistry reasoning & Text & \textcolor{unspecified}{N/A} & \textcolor{unspecified}{N/A} & Multiple chemistry tasks \\
  && ChemLLMBench \cite{guo2023can} & \textcolor{collection}{Varies by task} & Multiple tasks & Text & Multiple & Poor vs Chemformer & Molecular reasoning \\
  && SMol-Instruct \cite{yu2024llasmol} & \textcolor{quantified}{3M+ pairs} & Instruction QA & Text & Multiple & SOTA & Chemical Reaction \\
  \cmidrule{2-9}
  &\multirow{1}{*}{Molecular \&} & MaScQA \cite{zaki2023mascqa} & \textcolor{quantified}{650 QA} & QA & Text & GPT4 & ~62\% Accuracy & Materials Knowledge \\
  & \multirow{1}{*}{Materials Science}& LLM4Mat-Bench \cite{rubungo2024llm4mat} & \textcolor{quantified}{1.9M pairs} & Classification & Multiple & Multiple & \textcolor{unspecified}{N/A} & Property Prediction \\
  \cmidrule{2-9}
  & \multirow{2}{*}{Applied Chemistry}
  & PRESTO \cite{cao2024presto} & \textcolor{quantified}{3M samples} & Multiple tasks & Graphs \& Text & Multiple & SOTA & Molecule-text Modeling \\
  && DrugLLM \cite{liu2024drugllm} & \textcolor{unspecified}{N/A} & Generation & GMR & \textcolor{unspecified}{N/A} & \textcolor{unspecified}{N/A} & Drug design \\
  \cmidrule[\heavyrulewidth]{1-9}
  Environmental & \multirow{4}{*}{Climate \& Weather}
  & WeatherBench 2 \cite{rasp2024weatherbench} & \textcolor{quantified}{1-14 day} & Forecasting & Multivariate data & \textcolor{unspecified}{N/A} & \textcolor{unspecified}{N/A} & Medium-range forecasting \\
  Science && ClimateIQA \cite{chen2024vision} & \textcolor{quantified}{254,040 VQA} & VQA & Image + Text & Multiple & SOTA & Weather Event Detection \\
  && WeatherQA \cite{ma2024weatherqa} & \textcolor{quantified}{8000 VQA} & VQA & Image + Text & Multiple & Needs Improvement & Weather Event Detection \\
  && CLLMate \cite{li2024cllmate} & \textcolor{quantified}{26,156 QA} & Forecast & Multiple & Multiple & Needs Improvement & Event Forecasting \\
  \cmidrule{2-9}
  & \multirow{2}{*}{Disaster Response}
  & FFD-IQA \cite{sun2023unleashing} & \textcolor{quantified}{22,422 questions} & VQA & Image + Text & Multiple & Needs Improvement & Safety assessment \\
  && DisasterQA \cite{rawat2024disasterqa} & \textcolor{quantified}{707 questions} & MCQ & Text & GPT-4o & 85.78\% & Damage response \\
  \cmidrule{2-9}
  & \multirow{1}{*}{Pollution Monitoring}
  & VayuBuddy \cite{patel2024vayubuddy} & \textcolor{quantified}{7 years data} & Code Generation & Text & Multiple & \textcolor{unspecified}{N/A} & Sensor Data Analysis \\
  \cmidrule{2-9}
  & \multirow{2}{*}{Biodiversity \& Ecology}
  & Species-800 \cite{pafilis2013species} & \textcolor{quantified}{800 abstracts} & Classification & Text & \textcolor{unspecified}{N/A} & \textcolor{unspecified}{N/A} & NER \\
  && BiodivNERE \cite{abdelmageed2022biodivnere} & \textcolor{quantified}{2057 sentences} & Classification & Text & \textcolor{unspecified}{N/A} & \textcolor{unspecified}{N/A} & NER \\
  \cmidrule[\heavyrulewidth]{1-9}
  \end{tabular}%
  }
  
  \end{table*}

\subsection{Geography \& Remote Sensing}
The field of geography and remote sensing has seen remarkable advancements in the application of LLMs, particularly in processing and analyzing Earth observation data, climate patterns, agricultural systems, and complex geospatial information. Recent benchmarks demonstrate significant progress in multimodal capabilities, temporal reasoning, and domain-specific applications.

\subsubsection{Earth Observation}
Earth observation has emerged as a critical application domain for multimodal language models, with several significant benchmarks demonstrating the potential of these models in processing satellite and aerial imagery. The field has seen rapid advancement in both dataset development and model capabilities, particularly in temporal analysis and spatial reasoning tasks.

TEOChat \cite{irvin_teochat_2024} represents a significant breakthrough as the first vision-language assistant designed specifically for temporal satellite imagery analysis. The benchmark introduces TEOChatlas, a comprehensive dataset of 245,210 temporal examples spanning diverse tasks requiring both spatial and temporal reasoning. TEOChat demonstrates superior performance across seven task categories including temporal scene classification (75.1\% accuracy on fMoW RGB), change detection, and question answering, substantially outperforming previous vision-language models and even rivaling specialist models trained for specific tasks. The model's ability to process image sequences enables sophisticated analysis of dynamic phenomena such as disaster impacts and urban development, while its strong zero-shot generalization capabilities highlight the potential of multimodal approaches to enhance earth observation analysis for applications requiring temporal understanding.

EarthNets \cite{xiong_earthnets_2024} addresses the fragmentation in earth observation datasets by providing a comprehensive platform and benchmark for evaluating AI models on remote sensing data. The project systematically reviews and analyzes over 500 publicly available earth observation datasets spanning diverse domains including agriculture, land use, disaster monitoring, scene understanding, and climate change. What distinguishes EarthNets is its approach to dataset analysis across four dimensions: volume, resolution distributions, research domains, and inter-dataset correlations. The platform facilitates standardized evaluation of deep learning methods through unified dataset libraries and processing pipelines, effectively bridging the gap between remote sensing and machine learning communities. By enabling fair and consistent model comparisons, EarthNets represents a crucial step toward developing more robust foundation models for earth observation applications.

VLEO-Bench~\cite{zhang_good_2024} presents a comprehensive benchmark for evaluating Vision-Language Models (VLMs) on Earth observation data, with a particular focus on GPT-4V's capabilities. The benchmark assesses three key areas: scene understanding, localization and counting, and change detection, covering diverse applications such as urban monitoring, disaster relief, land use, and conservation. While GPT-4V demonstrates strong performance in open-ended tasks like location understanding and image captioning, it shows limitations in spatial reasoning tasks such as object localization and counting. The benchmark includes various datasets like RSICD for captioning, BigEarthNet for land cover classification, and xBD for change detection, providing a systematic evaluation framework for VLMs in Earth observation applications. This work highlights both the potential and current limitations of VLMs in processing satellite and aerial imagery, offering valuable insights for future model development in this domain.


The Earth Observation domain demonstrates significant progress in leveraging LLMs for geospatial and remote sensing tasks. Models in this field have showcased their ability to process diverse data types, such as satellite imagery and temporal sequences, providing actionable insights for tasks like environmental monitoring and urban planning. A notable trend is the increasing focus on temporal analysis, where LLMs are used to track dynamic changes over time. Despite these successes, the inherent complexity of geospatial data, including its high dimensionality and the need for precision, often limits the accuracy of existing models. Annotation quality remains another critical bottleneck; while automated annotations reduce costs and time, they may not capture subtle domain-specific nuances, leading to potential inaccuracies. The lack of unified evaluation metrics across geospatial tasks hinders consistent benchmarking and comparison of model performance.

\subsubsection{Climate \& Environment}
STBench \cite{li_stbench_2024} introduces a comprehensive framework for evaluating large language models' spatio-temporal analysis capabilities, particularly relevant for climate and environmental applications. The benchmark assesses four key dimensions: knowledge comprehension, spatio-temporal reasoning, accurate computation, and downstream applications, using over 60,000 question-answer pairs across 13 tasks.

The evaluation of 13 state-of-the-art LLMs, including GPT-4o and Gemma, reveals strong performance in knowledge comprehension and spatio-temporal reasoning tasks, with GPT-4o achieving 79.26\% accuracy on POI Category Recognition. However, models face challenges in accurate computation tasks. Advanced prompting techniques like in-context learning and chain-of-thought prompting show significant improvements, with ChatGPT's accuracy on POI Identification increasing from 58.64\% to 76.30\%.

While STBench's comprehensive evaluation framework is valuable for climate research, the lower performance on computation tasks suggests current LLMs need further development for precise environmental data analysis. The benchmark's open-source availability facilitates its adoption for climate and environmental research applications.

\subsubsection{Agricultural Science}
AgriLLM \cite{sapkota_multi-modal_2024} presents a comprehensive review of multi-modal large language models in agriculture, addressing 11 key research questions (4 general and 7 agriculture-focused) to understand their capabilities, challenges, and future directions. The survey systematically analyzes over 200 papers from multiple scientific databases, focusing on the integration of LLMs in agricultural applications.

The review highlights that multi-modal LLMs significantly enhance agricultural decision-making and image processing efficiency, particularly in areas such as crop monitoring, pest management, and yield prediction. These models demonstrate superior capabilities compared to traditional ML and DL methods, offering improved contextual understanding and predictive analytics. However, challenges remain in data privacy, computational costs, and the need for extensive domain-specific training data.


\subsubsection{Geospatial Analysis}
GEOBench-VLM \cite{danish_geobench-vlm_2024} introduces a comprehensive benchmark suite specifically designed for evaluating vision-language models (VLMs) on geospatial tasks, featuring over 10,000 manually verified instructions across 31 fine-grained tasks categorized into 8 broad categories including scene understanding, object counting, localization, segmentation, and temporal analysis. The benchmark addresses key challenges in geospatial data analysis such as temporal change detection, large-scale object counting, and tiny object detection, while employing multiple-choice questions to ensure objective and automated evaluation. Evaluation of 13 state-of-the-art VLMs reveals that while existing models show potential, they face significant challenges in geospatial tasks, with the best-performing LLaVA-OneVision achieving only 41.7\% accuracy on MCQs, slightly outperforming GPT-4o but still highlighting substantial room for improvement in geospatial-specific capabilities.

Roberts et al.~\cite{roberts_charting_2024} explores the geographic and geospatial capabilities of multimodal large language models (MLLMs), particularly focusing on GPT-4V's performance across various visual tasks. The study reveals that while GPT-4V demonstrates strong performance in recognizing fine details and reasoning about geographic features, it faces challenges in precise localization tasks. The evaluation of multiple MLLMs shows that model performance varies significantly by task type, with Qwen-VL and LLaVA-1.5 often outperforming GPT-4V in localization tasks, while GPT-4V excels in broader geographic understanding and reasoning. The study also highlights the models' tendency to struggle with multi-object images and their varying susceptibility to output format constraints.

RSUniVLM \cite{liu_rsunivlm_2024} presents a unified vision-language model specifically designed for remote sensing applications, featuring a novel Granularity-oriented Mixture of Experts architecture that enables comprehensive vision understanding across multiple granularity levels while maintaining a compact 1 billion parameter size. The model demonstrates strong performance across image-level tasks (captioning and visual question answering), region-level tasks (visual grounding and referring expression generation), and pixel-level tasks (semantic segmentation), while also supporting multi-image analysis for change detection and captioning. RSUniVLM's unified approach addresses the limitations of existing models in handling fine-grained pixel-level interpretation, which is crucial for applications like land-cover mapping and environmental protection.

INS-MMBench \cite{lin_ins-mmbench_2024} introduces the first comprehensive benchmark for evaluating large vision-language models (LVLMs) in the insurance domain, with particular relevance to geospatial analysis through its coverage of property and agricultural insurance tasks. The benchmark comprises 2.2K multiple-choice questions across 12 meta-tasks and 22 fundamental tasks, systematically organized through a bottom-up hierarchical task definition methodology. Evaluation of 10 LVLMs, including both closed-source (GPT-4o, GPT-4V) and open-source models (LLaVA, BLIP-2), reveals significant performance variations across insurance types, with GPT-4o achieving the highest score of 72.91/100. The benchmark's structured approach to task definition and evaluation provides valuable insights into LVLMs' capabilities in processing geospatial data for insurance applications, while highlighting challenges in domain-specific knowledge and perception accuracy.

\subsection{Physical \& Chemical Sciences}
The physical and chemical sciences present unique challenges for evaluating large language models, requiring deep domain knowledge, multi-step reasoning, and the ability to work with diverse representations including mathematical equations, chemical structures, and scientific diagrams. Recent benchmarks assess both foundational understanding and specialized capabilities across physics, chemistry, and interdisciplinary domains.

\subsubsection{Fundamental Science} 
One of the earlier benchmark datasets designed to evaluate the world knowledge and problem-solving abilities of large language models (LLMs) was MMLU (Massive Multitask Language Understanding) \cite{hendrycks2020measuring}. This dataset spans a wide range of topics, including foundational subjects in chemistry and physics such as mechanics and electromagnetism. It consists of multiple-choice questions sourced from high school and undergraduate exams. While MMLU has certain limitations typical of early benchmarks, it remains widely used within the research community and continues to evolve through iterative refinements \cite{gema2024we}.

Building on this, ScienceQA \cite{lu2022learn} was introduced as a large-scale multimodal benchmark, comprising approximately 21,000 multiple-choice questions across various science disciplines. Each question is annotated with supporting lectures, contextual information, and detailed explanations. A notable contribution of ScienceQA is its integration of chain-of-thought (CoT) reasoning, where the provided explanations simulate multi-step reasoning paths. This approach has been shown to improve LLM performance on science question answering tasks. The dataset is curated from elementary and high school curricula, and a significant portion of the questions are accompanied by images, enabling experiments that combine textual and visual modalities.

More recent benchmarks have focused on increasing the difficulty of evaluation tasks to better assess the capabilities of advanced models. One notable example is GPQA \cite{rein2024gpqa}, a highly challenging benchmark comprising 448 multiple-choice questions authored by domain experts in biology, physics, and chemistry. The difficulty of the dataset is underscored by the performance gap between experts and non-experts: even individuals with or pursuing PhDs in relevant fields achieve only around 65\% accuracy, while highly skilled non-expert annotators reach just 34\%—despite having unlimited web access and spending over 30 minutes per question. These ``Google-proof'' questions are designed to resist superficial search-based strategies. At the time of release, state-of-the-art models also struggled, with a GPT-4-based baseline achieving only 39\% accuracy. The authors emphasize the need for improved and scalable oversight mechanisms to match the increasing capabilities of large models.

Another recent addition is VisScience \cite{jiang2024visscience}, a multimodal benchmark specifically developed to evaluate the visual reasoning capabilities of MLLMs across scientific domains. The benchmark includes 3,000 K-12 level questions that integrate textual and visual modalities, spanning 21 science-related subjects and categorized into five levels of difficulty. Notably, VisScience highlights performance disparities across subjects and models. For instance, Claude 3.5-Sonnet achieved 53.4\% accuracy in mathematics, GPT-4o reached 38.2\% in physics, and Gemini 1.5 Pro scored 47.0\% in chemistry. These results suggest that even leading models face substantial challenges in visual scientific understanding, underscoring the need for further advancements in multimodal reasoning capabilities tailored to the diverse demands of STEM disciplines.

Meanwhile, IsoBench \cite{fu2024isobench} introduces a novel benchmarking framework to evaluate Multimodal Foundation Models on isomorphic representations—distinct formats conveying the same underlying information. A key focus of IsoBench is on scientific domains, particularly physics and chemistry, with 75 questions from each discipline. Each question is presented in multiple modalities, including visual (e.g., diagrams), textual (descriptive captions), and mathematical representations. This setup allows for a fine-grained analysis of how the form of input affects model performance. IsoBench reveals that current models exhibit a strong bias toward textual inputs. Notably, when evaluated across all IsoBench problems, Claude 3 Opus performs 28.7 percentage points worse when given images instead of equivalent textual descriptions. To mitigate such disparities, the authors propose prompting strategies—IsoCombination and IsoScratchPad—which encourage models to reason across, and translate between, different representations. These methods show measurable improvements, suggesting that aligning multimodal reasoning pathways can enhance model robustness. The dataset construction process is meticulous. For the visual modality, each question includes a figure paired with a textual prompt and multiple-choice options. The corresponding isomorphic textual input was manually annotated: an author provided a precise description of each figure, intentionally avoiding any additional inference beyond the content depicted. A portion of the IsoBench questions are adapted from existing benchmarks, including ScienceQA \cite{lu2022learn}, ensuring continuity and relevance with prior multimodal evaluation efforts.

Focusing more specifically on the domain of physics, MM-PhyQA \cite{anand2024mm} was developed to benchmark multi-step reasoning in high school-level physics tasks under a multimodal setting. The dataset includes physics questions accompanied by relevant visual inputs—such as diagrams, graphs, or circuit illustrations—and is designed to evaluate models' ability to perform complex reasoning that goes beyond surface-level pattern recognition. To support more advanced inference, the authors also release a CoT-Prompting variant, which incorporates exemplar questions during training to encourage chain-of-thought reasoning. The study provides an in-depth analysis of several factors that impact model performance: the inclusion of additional modalities beyond text, the use of CoT prompting techniques, and the effects of fine-tuning LLMs and LMMs for domain-specific tasks like physics question answering. One of the key contributions is the introduction of Multi-Image Chain-of-Thought (MI-CoT) prompting—a method that enables models to reason over multiple images alongside textual input, thereby more closely mirroring the way humans interpret and solve visual science problems.

For inference, the authors employed zero-shot GPT-4 for baseline predictions and evaluated LLaVA and LLaVA-1.5 models, both of which were fine-tuned on MM-PhyQA to assess performance gains from domain adaptation. For text-only evaluation, they tested both base and fine-tuned versions of Mistral-7B and LLaMA2-7B, comparing results across different prompting and fine-tuning strategies. Among all evaluated approaches, MI-CoT prompting combined with a fine-tuned LLaVA-1.5 13B model achieved the strongest performance, attaining the highest accuracy of 71.65\% on the test set and outperforming other models across most evaluation metrics. 

The field of chemistry and its related sub-disciplines have also witnessed the emergence of dedicated benchmarks aimed at evaluating LLMs' scientific reasoning and domain expertise. One such benchmark, ChemBench \cite{mirza2024large}, introduces a comprehensive and automated framework designed to assess the chemical knowledge and reasoning capabilities of state-of-the-art LLMs, comparing their outputs directly against those of trained chemists. The benchmark consists of 2,700 question-answer pairs, with each question annotated according to its topic, required skill (e.g., reasoning, calculation, factual knowledge, or intuition), and difficulty level. To establish human baselines, the authors developed a web-based interface and conducted a survey involving domain experts in chemistry. They then evaluated a range of both open- and closed-source LLMs, discovering that top-performing models, in aggregate, outperformed the best-performing human chemists in their study. However, this performance came with caveats: the models were prone to overconfident answers, particularly on simpler or foundational tasks, where basic errors were still prevalent.

Building on the momentum of multimodal reasoning in the chemical sciences, ChemQA \cite{chemQA2024} introduces a comprehensive multimodal question-answering dataset designed to evaluate the chemistry understanding of LLMs and MLLMs across a range of structured tasks. Inspired by benchmarks such as IsoBench \cite{fu2024isobench} and ChemLLMBench \cite{guo2023can}, ChemQA targets practical chemistry scenarios that require both symbolic reasoning and visual interpretation. The dataset includes five types of tasks: identifying atomic composition in organic molecules, calculating molecular weights, converting between SMILES strings and IUPAC names, generating and editing textual descriptions of molecules, and planning retrosynthetic pathways. These tasks collectively represent key challenges in computational chemistry and molecular design. ChemQA consists of approximately 85,000 examples, split into training, validation, and testing sets, and is available through Hugging Face.

ChemLLMBench \cite{guo2023can} offers a broad and systematic evaluation of large language models across a diverse set of real-world chemistry tasks. Designed to assess LLM capabilities in practical chemical reasoning, the benchmark spans eight task types, including bidirectional name conversion between IUPAC and SMILES, molecular property prediction (e.g., solubility and toxicity), reaction yield classification, forward reaction prediction, retrosynthesis planning, molecule captioning, reagent selection, and text-driven molecule design. It draws upon widely used datasets such as BBBP, Tox21 \cite{richard2020tox21}, PubChem \cite{kim2016pubchem}, USPTO \cite{jin2017predicting}, and ChEBI \cite{edwards2021text2mol} to construct its task-specific benchmarks. The scale of data varies depending on the task, ranging from thousands of molecules to reactions, and includes a mix of input modalities such as textual descriptions, SMILES strings, chemical formulas, and in some cases molecular graphs or images. Tasks cover both classification and generative formats. In their evaluation of five prominent LLMs—GPT-4, GPT-3.5, Davinci-003, LLaMA, and Galactica—GPT-4 consistently outperforms the others. However, the study finds that while LLMs perform strongly in classification-based or descriptive tasks like molecule captioning, they struggle with generative tasks that require deeper chemical understanding, such as reaction prediction and molecule synthesis.

SMol-Instruct \cite{yu2024llasmol} is a large-scale, high-quality dataset specifically designed for instruction tuning in the chemistry domain. It encompasses over three million samples across 14 carefully curated chemistry tasks, providing a robust foundation for training and evaluating large language models in a wide range of chemical reasoning challenges. The tasks span several categories, including name conversion between IUPAC, molecular formulas, and SMILES strings; molecule description through captioning and generation; property prediction across datasets like ESOL, LIPO, BBBP, ClinTox, HIV, and SIDER; as well as reaction tasks such as forward synthesis and retrosynthesis. To assess the effectiveness of instruction tuning, the authors fine-tune several open-source LLMs under the name LlaSMol, identifying Mistral as the most effective base model for chemistry tasks. SMol-Instruct's scale is significant, containing 3.3 million samples involving 1.6 million unique molecules, selected to reflect a broad diversity in molecular size, structure, and property space. The dataset undergoes rigorous quality control, including the removal of low-quality or erroneous samples, careful data partitioning, and the canonicalization of SMILES representations to ensure consistency and reliability across tasks.

\subsubsection{Molecular \& Materials Science} 
MaScQA \cite{zaki2023mascqa} (Materials Science Question Answering) is a benchmark specifically designed to evaluate large language models on undergraduate-level materials science, a field that bridges physics and chemistry. The dataset contains 650 questions categorized by subtopics such as material structure, properties, processing, and phase behavior, with a focus on assessing knowledge of concepts like crystallography and phase diagrams. The authors evaluate GPT-3.5 and GPT-4 using both zero-shot and chain-of-thought prompting, finding that GPT-4 performs best with an accuracy of around 62\%. Notably, unlike in many other domains, chain-of-thought prompting did not yield a significant improvement in performance, suggesting that conventional prompting techniques may have limited effectiveness in highly specialized scientific fields like materials science.

Similarly, LLM4Mat-Bench \cite{rubungo2024llm4mat} represents an extensive benchmark to date for assessing the capability of large language models in predicting properties of crystalline materials. The dataset comprises approximately 1.9 million crystal structures curated from ten publicly available materials science databases and spans 45 distinct material properties. It supports multiple input modalities, including crystal composition, crystallographic information files (CIFs), and textual descriptions of crystal structures, encompassing 4.7 million, 615.5 million, and 3.1 billion tokens, respectively. The benchmark is used to fine-tune and evaluate various models, such as LLM-Prop and MatBERT, across these input types. Results reveal that general-purpose LLMs often fall short in accurately predicting materials properties, emphasizing the need for domain-adapted, instruction-tuned models specifically designed for materials science applications.

\subsubsection{Applied Chemistry} 
A recent effort aimed at advancing multimodal modeling in synthetic chemistry is PRESTO \cite{cao2024presto} (Progressive Pretraining Enhances Synthetic Chemistry Outcomes), a framework designed to bridge the molecule-text modality gap in reaction-centric tasks. Synthetic chemistry, which focuses on designing and executing chemical reactions to produce compounds with specific properties, presents unique challenges for multimodal learning due to the complexity of reaction procedures and molecular representations. PRESTO addresses this by integrating a large-scale dataset of approximately 3 million samples, consisting of synthetic procedure descriptions and molecule name conversions. It provides a structured benchmark for evaluating various pretraining strategies and dataset configurations. The framework supports a range of tasks critical to synthesis, including reaction prediction, reaction condition prediction, reagent selection, reaction type classification, and yield regression. Through progressive pretraining techniques that promote cross-modal alignment and multi-graph understanding, PRESTO demonstrates notable improvements in the performance of multimodal LLMs across these tasks, offering a robust foundation for future work in automated reaction planning and chemical synthesis modeling.

In the domain of drug discovery, the DrugLLM \cite{liu2024drugllm} dataset introduces a large-scale, specialized resource designed to train LLMs for molecule generation and optimization. At its core is the Group-based Molecular Representation (GMR), which sequences molecular modifications aimed at improving specific chemical properties. By learning to predict successive molecular edits, DrugLLM models the process of rational drug design. The dataset draws from two well-established sources, ZINC~\cite{irwin2020zinc20} and ChEMBL \cite{davies2015chembl}, transforming tabular molecular data into structured ``modification paragraphs''—narratives of molecular changes associated with over 10,000 different properties or activities. In total, the dataset includes over 25 million paragraphs and 200 million training samples, making it one of the largest resources of its kind. Empirical results show that DrugLLM exhibits strong few-shot generation capabilities, effectively proposing novel molecules with desired properties using limited input examples, positioning it as a promising tool for accelerating drug design through language-based modeling.

\subsection{Environmental Science}
Environmental science presents unique challenges for large language models, requiring them to reason about complex natural systems, process multimodal data from diverse sensors and satellites, and support critical decision-making in areas like climate change, disaster response, and ecological conservation. This section examines how LLMs and MLLMs are being evaluated and applied across key environmental domains including weather forecasting, disaster management, pollution monitoring, and biodiversity assessment.

\subsubsection{Climate \& Weather}
WeatherBench 2 \cite{rasp2024weatherbench} is an update to the global, medium-range (1--14 day) weather forecasting benchmark called WeatherBench \cite{rasp2020weatherbench}, designed with the aim to accelerate progress in data-driven weather modeling. WeatherBench 2 consists of an open-source evaluation framework, publicly available training, ground truth and baseline data as well as a continuously updated website with the latest metrics and state-of-the-art models. Studies have leveraged this data and used LLMs in weather forecasting. For example, CLIMATELLM \cite{li2025climatellm} captures spatiotemporal dependencies via a cross-temporal and cross-spatial collaborative modeling framework that integrates Fourier-based frequency decomposition with Large Language Models (LLMs) to strengthen spatial and temporal modeling.

ClimateIQA \cite{chen2024vision} a first meteorological VQA dataset, which includes 8,760 wind gust heatmaps and 254,040 question-answer pairs covering four question types, both generated from the latest climate reanalysis data. The researchers redefines Extreme Weather Events Detection (EWED) by framing it as a Visual Question Answering (VQA) problem, thereby introducing a more precise and automated solution. Leveraging Vision-Language Models (VLM) to simultaneously process visual and textual data, we offer an effective aid to enhance the analysis process of weather heatmaps.  In the creation process, images were processed using SPOT to extract color contours and representative point coordinates. The extracted data were integrated into geographic knowledge bases to retrieve location-specific information. These data, including location, coordinates, and wind speed, were then input into predefined question-and-answer templates, resulting in the generation of 254,040 question-answer pairs. 

WeatherQA \cite{ma2024weatherqa} is a multimodal dataset designed for machines to reason about complex combinations of weather parameters and predict severe weather in real-world scenarios. Collected and processed from the NOAA Storm Prediction Center, the dataset includes 8,511 (multi-image, text) pairs of weather data from severe convective and winter storms together with expert analyses. Each pair contains rich information crucial for forecasting—the images describe the ingredients capturing environmental instability, surface observations, and radar reflectivity, and the text contains in-depth forecast analyses written by human experts. They also evaluate many vision language models (VLMs) in two tasks: (1) multi-choice QA for predicting the affected area and (2) classification of the development potential of severe convection. Results show a substantial gap between the strongest VLM, GPT4o, and human reasoning. They suggest that better training and data integration are necessary to bridge this gap.
 
\subsubsection{Disaster Response}
Natural disasters pose significant threats to human life and infrastructure. Multimodal Large Language Models (MLLMs) show promise in enhancing disaster response by processing both textual and visual data to support emergency management. However, given the high-stakes nature of disaster scenarios, rigorous evaluation of MLLMs' capabilities is essential before deployment. Recent benchmarks assess both textual reasoning and visual understanding in disaster contexts, focusing on real-world applicability.

The FFD-IQA (Freestyle Flood Disaster Image Question Answering) benchmark \cite{sun2023unleashing} addresses the challenge of visual question answering in flood disaster scenarios. The dataset comprises 2,058 images paired with 22,422 question-answer pairs spanning free-form, multiple-choice, and yes-no question formats. Unlike previous disaster-focused VQA benchmarks, FFD-IQA expands beyond closed-answer spaces to allow for more nuanced responses about disaster conditions. The authors also introduce a Zero-shot VQA for Flood Disaster Damage Assessment (ZFDDA) model, which leverages chain-of-thought (CoT) demonstrations to enhance reasoning capabilities without requiring pre-training on disaster-specific data. Their experimental results demonstrate that implementing CoT prompts significantly improves model accuracy, particularly for complex questions requiring multi-step reasoning. The benchmark specifically focuses on evaluating safety assessment capabilities—determining whether individuals are in danger and whether emergency services are responding adequately—making it particularly relevant for real-time disaster management applications.

Complementing the visual understanding focus of FFD-IQA, DisasterQA \cite{rawat2024disasterqa} provides a comprehensive text-based benchmark for evaluating LLMs' knowledge of disaster response protocols and best practices. The dataset consists of 707 multiple-choice questions curated from six reliable online sources, covering a wide range of disaster response topics. The authors evaluated five leading LLMs (GPT-3.5 Turbo, GPT-4 Turbo, GPT-4o, Llama 3.1, and Gemini) using four different prompting methods: standard prompting, directional stimulus prompting, chain-of-thought prompting, and few-shot prompting. Their results indicate that GPT-4o achieved the highest accuracy of 85.78\% when using chain-of-thought prompting, demonstrating promising capabilities but also highlighting room for improvement before these models can be reliably deployed in critical disaster response scenarios. Notably, the study revealed that certain types of disaster response questions were consistently answered incorrectly across all models, suggesting fundamental knowledge gaps that need to be addressed. The benchmark emphasizes the importance of not just accuracy but also appropriate confidence calibration, as overconfident yet incorrect responses could be particularly harmful in emergency situations.

\subsubsection{Pollution Monitoring}
The effective monitoring of environmental pollutants represents a critical application domain for multimodal language models, with direct implications for public health and environmental policy. Air pollution alone accounts for approximately 6.7 million deaths annually, making timely access to accurate pollutant data essential for mitigating exposure risks. While government agencies routinely collect extensive environmental sensor data, there exists a significant barrier between this raw information and various stakeholders who could benefit from it. VayuBuddy \cite{patel2024vayubuddy} addresses this challenge through an LLM-powered chatbot system that bridges the gap between technical sensor data and diverse stakeholders. Using seven years of air quality data from Indian government sensors, the benchmark evaluates models' ability to generate Python code that analyzes structured sensor data in response to natural language queries about air pollution. The system assesses various LLMs including Llama, Mixtral, Codestral, and Gemma across 45 diverse question-answer pairs. While current models show promising performance in generating accurate code for data analysis, significant challenges remain in handling complex queries and ensuring consistent performance across different types of environmental analysis tasks. The benchmark demonstrates the potential of MLLMs to democratize access to environmental data by enabling stakeholders to derive meaningful insights without specialized technical expertise.

\subsubsection{Biodiversity \& Ecology}
The monitoring and analysis of biodiversity data is crucial for ecological research and conservation. With rapidly growing scientific literature in this field, automated tools are needed to extract information about species and their relationships from text. Named Entity Recognition (NER) and Relation Extraction (RE) face challenges due to complex taxonomic nomenclature and specialized vocabulary. MLLMs show promise in transforming biodiversity informatics by automating information extraction from scientific literature and enabling comprehensive ecological analysis.

The Species-800 (S800) corpus \cite{pafilis2013species} provides a gold-standard dataset for evaluating named entity recognition of species and other taxonomic mentions in biomedical text. The corpus comprises 800 manually annotated PubMed abstracts spanning eight taxonomic categories, including viruses, bacteria, plants, fungi, and various animal groups. Unlike previous taxonomic datasets, S800 represents a diverse range of organisms and publication sources, offering insights into which types of taxonomic names are particularly challenging for models to identify correctly. The benchmark was developed to support the SPECIES tagger, a dictionary-based approach to taxonomic NER that combines the NCBI Taxonomy with expanded synonyms and variant forms. The authors' evaluation demonstrates that their approach achieves comparable accuracy to existing tools while being more than an order of magnitude faster, making it suitable for processing large document collections. The S800 corpus reveals category-specific performance differences, with virus names posing particular challenges compared to other taxonomic groups. This benchmark not only supports the development of more efficient taxonomic NER tools but also facilitates the creation of resources like ORGANISMS, a web-based platform that provides access to pre-computed species tagging results for the entire Medline database.

Expanding beyond species identification, the BiodivNERE dataset \cite{abdelmageed2022biodivnere} provides a more comprehensive benchmark for both named entity recognition and relation extraction in biodiversity literature. This gold-standard corpus includes 2,057 manually annotated sentences covering six entity classes and 17 relation types derived from biodiversity research questions and ontologies. The dataset was constructed through a rigorous process involving biodiversity experts, with annotation guidelines designed to ensure consistency and accuracy. The annotation schema incorporates a wide range of entities relevant to biodiversity research, including not just organisms but also habitats, environmental features, materials, processes, and qualities. The relation types capture important ecological and evolutionary connections between these entities, such as ``lives\_in'' relationships between organisms and habitats or ``affects'' relationships between processes and organisms. By providing benchmarks for both NER and RE tasks, BiodivNERE enables the development of more sophisticated tools for analyzing the rapidly growing biodiversity literature. The authors' evaluation reveals high inter-annotator agreement and provides detailed statistics on entity and relation distributions, offering valuable insights into the challenges of information extraction in this domain. Together with S800, these benchmarks highlight the complexity of biodiversity information extraction and provide essential resources for training and evaluating MLLMs in ecological applications.

\section{Technology}\label{sec:technology}
Technology domains represent critical application areas where MLLMs are transforming capabilities across industries. These domains leverage multimodal reasoning to enhance perception, decision-making, and human-machine interaction in complex technical environments. An overview of all the relevant benchmarks is presented in Table~\ref{tbl:technology_discipline}.

\begin{table*}[!ht]
  \caption{Technology Discipline: Domain-Specific Benchmarks}\label{tbl:technology_discipline}
  \centering
  
  \resizebox{\fulllength}{!}{%
  \begin{tabular}{lllcccccc} 
  \cmidrule[\heavyrulewidth]{1-9}
  \textbf{Domain} & \textbf{Sub-domain} & \textbf{Benchmark} & \textbf{Scale} & \textbf{Task Type} & \textbf{Input Modality} & \textbf{Model} & \textbf{Performance} & \textbf{Key Focus} \\
  \cmidrule[\heavyrulewidth]{1-9}
  \multirow{12}{*}{Computer Vision \&} 
  & \multirow{4}{*}{Visual Understanding}
  & Rank2Tell~\cite{sachdeva2024rank2tell} & \textcolor{unspecified}{N/A} & Importance ranking & Multi-modal & \textcolor{unspecified}{N/A} & \textcolor{unspecified}{N/A} & Object importance \\
  \multirow{12}{*}{Autonomous Systems}&& NuInstruct~\cite{ding2024holistic} & \textcolor{quantified}{91K pairs} & Multi-view QA & Video + BEV & \textit{BEV-InMLLM} & +9\% & Holistic understanding \\
  && NuScenes-QA~\cite{qian_nuscenes-qa_2024} & \textcolor{quantified}{460K QA pairs} & Multi-modal VQA & Image + LiDAR & Multiple & \textcolor{unspecified}{N/A} & Multi-frame VQA \\
  && Cambrian-1~\cite{tong_cambrian-1_2024} & \textcolor{collection}{20+ encoders} & Vision-centric & Multi-modal & Cambrian-1 & SOTA & Visual grounding \\
  \cmidrule{2-9}
  & \multirow{8}{*}{Autonomous Driving}
  & HAZARD~\cite{zhou_hazard_2024} & \textcolor{quantified}{3 scenarios} & Decision making & Multi-modal & LLM agent & \textcolor{unspecified}{N/A} & Disaster response \\
  && LLM4Drive~\cite{yang_llm4drive_2024} & Survey paper & System review & Multi-modal & Multiple & \textcolor{unspecified}{N/A} & System architecture \\
  && WTS~\cite{kong_wts_2025} & \textcolor{quantified}{1.2k events} & Video analysis & Video + Text & VideoLLM & \textcolor{unspecified}{N/A} & Pedestrian safety \\
  && DRAMA~\cite{malla2023drama} & \textcolor{quantified}{17,785 scenarios} & Risk assessment & Video + Objects & \textcolor{unspecified}{N/A} & \textcolor{unspecified}{N/A} & Risk localization \\
  && GenAD~\cite{yang_generalized_2024} & \textcolor{quantified}{2000+ hours} & Video prediction & Video + Text & \textit{GenAD} & \textcolor{unspecified}{N/A} & Generalized prediction \\
  && Reason2Drive~\cite{nie_reason2drive_2025} & \textcolor{quantified}{600K+ pairs} & Chain reasoning & Video + Text & VLMs & \textcolor{unspecified}{N/A} & Interpretable reasoning \\
  && DriveLM~\cite{sima_drivelm_2025} & \textcolor{unspecified}{N/A} & Graph VQA & Multi-modal & \textit{DriveLM-Agent} & \textcolor{unspecified}{N/A} & End-to-end driving \\
  && NuPrompt~\cite{wu_language_2023} & \textcolor{quantified}{35,367 prompts} & Object tracking & Multi-view & \textit{PromptTrack} & \textcolor{unspecified}{N/A} & Language prompts \\
  \cmidrule[\heavyrulewidth]{1-9}
  \multirow{2}{*}{Robotics \&} 
  & \multirow{1}{*}{Robot Control}
  & MMRo~\cite{li2024mmro} & \textcolor{unspecified}{N/A} & Manufacturing & Multi-modal & Multiple & \textcolor{unspecified}{N/A} & Manufacturing \\
  \cmidrule{2-9}
  \multirow{1}{*}{Automation}& \multirow{1}{*}{Process Automation}
  & DesignQA~\cite{designqa} & \textcolor{unspecified}{N/A} & Design & Multi-modal & Multiple & \textcolor{unspecified}{N/A} & Design \\
  \cmidrule[\heavyrulewidth]{1-9}
 \multirow{10}{*}{Blockchain \&} 
  & \multirow{3}{*}{Smart Contract Analysis}
  & Web3Bugs~\cite{xiao2025logicmeetsmagicllms} & \textcolor{collection}{Multiple versions} & Security classification & Code (Solidity) & GPT-4 (prompted) & 60\% false-positive reduction & Bug detection \\
  \multirow{10}{*}{Cryptocurrency}& & LLM-SmartAudit~\cite{wei2024llmsmartauditadvancedsmartcontract} & \textcolor{collection}{Two datasets} & Multi-agent audit & Code + Documentation & \textit{LLM-SmartAudit} & Outperforms conventional tools & Comprehensive auditing \\
  & & ACFIX~\cite{zhang2024acfixguidingllmsmined} & \textcolor{quantified}{118 contracts} & Auto-repair & Code (Solidity) & GPT-4 + RBAC & 94.9\% fix rate & Patch generation \\
  \cmidrule{2-9}
  & \multirow{3}{*}{Market Analysis}
  & CryptoNews~\cite{Roumeliotis2024LLMsCrypto} & \textcolor{quantified}{3,200 articles} & Sentiment analysis & News text & FT-GPT-4 & 86.7\% accuracy & Market sentiment \\
  & & Ethereum Prices~\cite{makri2025ethereumpricepredictionemploying} & \textcolor{collection}{5 years daily} & Price forecasting & Time-series & GPT-2 (FT) & SOTA MSE & Short-term pred. \\
  & & LLM-Trading~\cite{wang2025exploringllmcryptocurrencytrading} & \textcolor{collection}{Multi-modal} & Strategy reasoning & Text + Indicators & GPT-4 agent & ↑PnL in sim. & Fact-subjectivity reasoning \\
  \cmidrule{2-9}
  & \multirow{1}{*}{Fraud Detection}
  & BLOCKGPT~\cite{he2025largelanguagemodelsblockchain} & \textcolor{quantified}{M txns} & Anomaly detection & Txn graph & \textit{BLOCKGPT} & 40\% attack detection & Real-time monitoring \\
  \cmidrule{2-9}
  & \multirow{1}{*}{Governance}
  & Crypto Legal Cases~\cite{trozze2024largelanguagemodelscryptocurrency} & \textcolor{collection}{SEC files} & Legal reasoning & Case text & GPT-3.5 & Mixed & Compliance assistance \\
  \cmidrule[\heavyrulewidth]{1-9}
  \end{tabular}%
  }
  
  \end{table*}

\subsection{Computer Vision \& Autonomous Systems}

Autonomous systems rely on computer vision to understand their environment and make quick decisions. They use tasks like object detection, scene segmentation, and predicting movement to stay aware and safe. With the rise of multimodal large language models (MLLMs), these systems can now combine images and text to improve reasoning, planning, and interaction with humans. 

Recent research shows that foundational computer vision tasks such as semantic segmentation, object tracking, and depth estimation are being reimagined for autonomous driving using grid-centric and multimodal approaches. Furthermore, by fusing LiDAR, video, and natural language, Multimodal Large Language Models enable interpretable, instruction-based decision-making for autonomous systems \cite{cui2024survey,shi2024med}. This section is divided into two key subdomains: Visual Understanding and Autonomous Driving, each highlighting a different facet of how computer vision integrates into autonomous systems through MLLMs.

\subsubsection{Visual Understanding}
Visual understanding is foundational for deploying Multimodal Large Language Models (MLLMs) in autonomous driving, where systems must interpret complex, dynamic scenes using visual, spatial, and temporal cues. While early benchmarks focused on general-purpose Visual Question Answering (VQA), recent efforts prioritize domain-specific reasoning, particularly concerning safety, planning, and contextual awareness. Emerging datasets now evaluate how effectively MLLMs can ground language in real-world driving scenes.

A key shift involves an emphasis on spatial-temporal reasoning. NuScenes-QA introduces a large-scale benchmark with over 460,000 question-answer pairs built from RGB and LiDAR inputs across multiple frames. It probes dynamic object tracking, spatial relationships, and causal inference in traffic scenarios~\cite{qian_nuscenes-qa_2024}. Building on this, NuInstruct adopts an instruction-driven format with 91,000 prompts covering 17 driving subtasks, incorporating multi-view video, temporal dynamics, and bird's-eye-view (BEV) representations. The benchmark includes the BEV-InMLLM model, which aligns BEV features with LLM reasoning to simulate human-like decision-making logic \cite{ding2024holistic}. Together, these benchmarks highlight a shift from static image-based VQA to structured, temporally-aware evaluations.

Another major advancement focuses on agent prioritization and explainability. Rank2Tell presents a multimodal dataset where models must not only identify traffic participants but also rank them by importance and provide natural language justifications. It evaluates whether models can reason about risk, attention, and intent, mirroring human driver cognition and interpretability needs \cite{sachdeva2024rank2tell}.

In addition, Cambrian-1 focuses on the internal visual representation quality of MLLMs rather than downstream performance. It converts classical computer vision tasks into VQA format and introduces the Spatial Vision Aggregator (SVA) to fuse high-resolution visual features into LLMs. Results show that visual grounding—rather than language reasoning—is often the limiting factor in MLLM performance, especially in tasks requiring spatial disambiguation or high-detail understanding \cite{tong2024ploutos}.

Collectively, these benchmarks reflect a trend toward more holistic and interpretable visual understanding spanning perception, spatial reasoning, and explanation. However, current models still struggle with BEV-based and multi-view reasoning, high-resolution perception, and agent prioritization. Evaluation protocols remain inconsistent, hindering cross-benchmark comparisons. Table~\ref{tbl:technology_discipline} summarizes the design goals and task types of each benchmark.

Going forward, unified evaluation strategies combining spatial fidelity, multi-agent reasoning, and instruction-based interaction are needed. Innovations like modular vision encoders, BEV fusion, and justification modules may be essential for real-world deployment of MLLMs in safety-critical driving environments.

\subsubsection{Autonomous Driving}
The integration of Multimodal Large Language Models (MLLMs) into autonomous driving systems has led to growing interest in evaluating how these models perceive, reason, and act within highly dynamic and safety-critical environments. Existing benchmarks explore a range of capabilities, from low-level perception to high-level reasoning and decision-making, offering diverse perspectives on how MLLMs interact with visual and textual driving data. Rather than focusing on single datasets or metrics, recent works emphasize the need for holistic evaluation frameworks that test the robustness, explainability, and generalization ability of MLLMs under real-world constraints.

Several benchmarks focus on fine-grained perception and the ability of MLLMs to generate semantically meaningful representations of complex traffic scenes. The WTS dataset offers a pedestrian-centric benchmark that combines multi-view videos with synchronized 2D/3D spatial data, textual event descriptions, and 3D gaze annotations to capture nuanced pedestrian and vehicle behaviors \cite{kong_wts_2025}. It introduces LLMScorer, a semantic similarity-based evaluation metric tailored to long, structured video captions, overcoming the limitations of traditional n-gram metrics. This work highlights the importance of understanding human behaviors in real-time, particularly when interpreting intent in dense, urban scenarios. Complementing this, the DRAMA dataset introduces a joint framework for risk localization and captioning, where annotated driving scenarios focus on identifying critical objects and explaining the reasoning behind risk predictions \cite{malla2023drama}. DRAMA pushes beyond classification by integrating visual question answering with natural language reasoning, providing both open-ended and structured evaluation formats. These datasets mark a shift from pure perception to context-aware safety reasoning, bridging the gap between detection and interpretable understanding.

A second group of benchmarks focuses more explicitly on causal reasoning and structured decision processes. The Reason2Drive benchmark offers over 600,000 video-text pairs that model the full driving pipeline—perception, prediction, and reasoning—through chain-of-thought annotations derived from datasets such as nuScenes, Waymo, and ONCE \cite{nie_reason2drive_2025}. It proposes a novel evaluation metric, ADRScore, which captures the logical quality of reasoning chains beyond surface-level textual similarity. Similarly, DriveLM introduces Graph Visual Question Answering (GVQA) as a proxy for evaluating object-centric reasoning in autonomous driving, incorporating perception, behavior, and planning questions linked through graph-structured dependencies~\cite{sima_drivelm_2025}. Unlike single-turn QA datasets, DriveLM supports multi-step logic grounded in object interactions and scenario evolution. Together, these benchmarks reveal a shift toward evaluating models not only on recognition accuracy but also on their ability to construct interpretable, stepwise justifications for driving decisions--an essential requirement for deployment in high-stakes applications.

Another set of works explores the use of natural language prompts to drive perception and planning. NuPrompt extends the nuScenes dataset with over 40,000 prompt-object trajectory pairs, allowing evaluation of MLLMs' ability to ground language instructions to multi-object, multi-view 3D scenes \cite{wu_language_2023}. Unlike earlier prompt-based vision tasks that focus on static images or individual objects, NuPrompt supports object tracking across frames and views, enabling evaluation of temporal and spatial grounding at scale. This benchmark introduces a new class of evaluation tasks that blend natural language interaction with multimodal tracking and planning. GenAD further expands this vision by introducing a generative video prediction model trained on more than 2,000 hours of driving footage, called OpenDV-2K \cite{yang2024editworld}. The model predicts future driving scenes given textual instructions and past observations, enabling zero-shot generalization to new environments, weather conditions, and sensor configurations. By treating driving as a predictive task, GenAD demonstrates that generative modeling can support both action planning and environment simulation across unseen conditions.

Beyond perception and planning, the HAZARD benchmark introduces dynamic environments such as fires, floods, and wind, challenging MLLMs to adapt to non-static, unfolding disasters~\cite{zhou_hazard_2024}. This simulation-based benchmark pushes the boundaries of embodied AI by requiring models to perform real-time decision-making in unstable settings. An LLM API is provided to convert raw visual and memory data into semantic text representations, allowing models to plan and act through natural language-based policies. This opens a new line of research into embodied reasoning under temporal stress and environmental uncertainty, areas often overlooked in conventional benchmarks. Finally, the LLM4Drive survey offers a meta-perspective on this evolving field, identifying key trends, limitations, and future opportunities in applying LLMs to autonomous systems~\cite{yang_llm4drive_2024}. It discusses how modular perception-prediction-planning pipelines are increasingly being replaced by unified, prompt-driven architectures capable of open-world reasoning, few-shot learning, and contextual adaptation. The survey also raises concerns about interpretability, sim-to-real generalization, and the lack of standardized evaluation metrics across tasks.

Together, these benchmarks form a comprehensive and multi-layered foundation for evaluating MLLMs in autonomous driving. While they vary in focus from pedestrian intent detection to graph-based reasoning and disaster-aware planning, they all highlight the need for interpretable, robust, and generalizable models. A clear research gap persists in unified evaluation protocols and cross-dataset comparability, particularly in reasoning accuracy and safety-critical contexts. As LLMs and MLLMs become integral to autonomous systems, future work must address these limitations by integrating standardized, high-stakes, and explainable evaluation practices that reflect real-world complexity and operational constraints.

\subsection{Robotics \& Automation}
Multimodal Large Language Models (MLLMs) are playing a bigger role in robotics and automation because they can combine language, vision, and reasoning. In robotics, they help plan and understand tasks, while in automation, they support decision-making using visual and language inputs. However, to properly evaluate their performance, we need strong, domain-specific tests that go beyond basic results. This section focuses on two key areas—robot control and process automation, each with different needs for perception, planning, and safety.

\subsubsection{Robot Control}
Recent efforts to assess the viability of MLLMs as ``brains'' for in-home robots have led to the development of the MMRo benchmark, which provides a structured evaluation of MLLMs across four robotics-specific dimensions: perception, task planning, visual reasoning, and safety \cite{li2024mmro}. Unlike traditional VQA benchmarks that focus on general knowledge, MMRo targets domain-relevant scenarios such as object manipulation, spatial awareness, and hazard recognition. Each task is framed through real-world or synthetic images and paired with both multiple-choice and open-ended questions, allowing fine-grained analysis of model behavior.

Experimental results reveal that while top-performing models like GPT-4V and Gemini-Pro excel in high-level reasoning tasks (e.g., object function recognition or plan sequencing), they struggle with basic perceptual capabilities such as recognizing shapes or identifying object materials. These deficits raise concerns about their reliability in real-world deployment. The benchmark also identifies significant inconsistencies across models in safety-related tasks. For example, correctly identifying when an object is hot or sharp is critical for safe manipulation and is a challenge for both commercial and open-source MLLMs. This indicates that while MLLMs may enable abstract reasoning, they currently lack the precision and robustness required for embodied control. The MMRo benchmark thus serves not only as an evaluation tool but also as a diagnostic framework, highlighting specific failure modes that must be addressed for MLLMs to serve as viable cognitive cores in robotics \cite{li2024mmro}.

\subsubsection{Process Automation}
In the broader field of automation, MLLMs are also being evaluated for tasks like Earth observation, disaster monitoring, and environmental assessment. A recent benchmark evaluates GPT-4V on high-resolution satellite imagery to test its performance on scene understanding, object counting, and change detection~\cite{zhang_good_2024}. The study focuses on zero-shot generalization—an essential capability for real-world automation—since models are not fine-tuned on remote sensing data. GPT-4V performs well on descriptive tasks such as image captioning and landmark recognition but shows limited effectiveness in spatially demanding tasks like counting buildings or identifying changes in flood-prone regions.

The benchmark also highlights challenges in grounding language to dense, structured data. For instance, while GPT-4V can describe scenes fluently, it often misinterprets spatial relationships and object quantities, shortcomings that would severely impact downstream decision-making in infrastructure or crisis management. Although promising as natural language front-ends for automated systems, MLLMs currently fall short of the precision required for scalable, real-world automation. The study emphasizes the need for better alignment between language models and geospatial reasoning modules to ensure safe and actionable outcomes.
\subsection{Blockchain \& Cryptocurrency}
Blockchain and cryptocurrency technologies present unique challenges for multimodal AI systems, requiring specialized understanding of code, finance, and decentralized systems. This section examines how MLLMs are being evaluated and applied across smart contract security, market analysis, fraud detection, and governance tasks.

\subsubsection{Brief Introduction}
The blockchain and cryptocurrency domain poses unique challenges and opportunities for applying large language models. Smart contracts underpin decentralized finance and applications, carrying billions in value and demanding high security. Likewise, cryptocurrency markets are heavily driven by public sentiment and complex token economies. LLMs have begun to serve as powerful tools in this domain, from auditing smart contract code to interpreting market sentiment, due to their human-like language and code understanding \cite{he2025largelanguagemodelsblockchain,xiao2025logicmeetsmagicllms}. This section surveys how both text-only and multimodal LLMs are being leveraged across blockchain security, crypto market analysis, and decentralized decision-making. We highlight key application areas – smart contract analysis, sentiment forecasting, fraud detection, and governance – and discuss emerging trends and challenges.

\subsubsection{Smart Contract Analysis and Auditing}
One critical application of LLMs in blockchain is \textbf{smart contract auditing}. Smart contracts are self-executing programs (often in Solidity) that manage digital assets, and vulnerabilities in them have led to major financial losses \cite{xiao2025logicmeetsmagicllms}. Traditional static analyzers and fuzzers detect some bugs but often miss complex logic flaws. LLMs offer a new angle by understanding code semantics and even natural-language documentation of contracts. Early studies using GPT-3.5/4 to detect Solidity bugs found encouraging recall but alarmingly high false-positive rates. For instance, one analysis noted that GPT-4-based detectors achieved relatively high recall on known vulnerability types but with precision trade-offs, especially on newer Solidity versions \cite{xiao2025logicmeetsmagicllms}. Recent research tackles these limits: carefully engineered prompts can cut false alarms by over 60\%, and fine-tuning or combining LLMs with symbolic analysis significantly boosts accuracy. Novel frameworks like \textbf{LLM-SmartAudit} employ multiple specialized LLM agents in a cooperative manner to audit code, outperforming all conventional tools on broad vulnerability benchmarks \cite{wei2024llmsmartauditadvancedsmartcontract}. These LLM-driven auditors not only catch standard bugs (reentrancy, arithmetic errors) but even complex logic or business rule violations that rule-based scanners overlooked. Beyond detection, LLMs have also been used to \textbf{repair vulnerabilities} by suggesting fixes in code. Early results show GPT-4 can propose patches for certain security bugs (e.g. access control flaws) when guided with the right context \cite{he2025largelanguagemodelsblockchain}. Likewise, developers are using code-aware LLMs (e.g. GitHub Copilot or ChatGPT) to \textit{generate} smart contracts and unit tests. This accelerates development but comes with the risk of the model introducing subtle bugs \cite{xiao2025logicmeetsmagicllms}. Overall, LLMs in smart contract analysis demonstrate \textbf{promising capabilities} in reasoning about code and security; the trend is toward hybrid approaches (LLM + static analysis + fuzzing) to mitigate errors. Key challenges remain in keeping LLMs up-to-date with evolving languages (e.g. Solidity changes that confused earlier models) and guaranteeing reliability to a level suitable for high-stakes financial code.

\subsubsection{Market Analysis and Sentiment Forecasting}
Cryptocurrency markets are notoriously volatile and driven by news, social media, and investor sentiment. LLMs have begun playing a role in \textbf{crypto market analysis} by extracting and forecasting sentiment signals from text. Financial NLP was an early success area for domain-specific language models (e.g. FinBERT for stock sentiment), and similar approaches are now applied to crypto \cite{shah2022when}. A recent study fine-tuned GPT-4 on \textbf{cryptocurrency news articles} to classify sentiment, achieving $\sim$86.7\% accuracy -- slightly surpassing finance-tuned BERT models (FinBERT $\sim$84.3\%) on a test set of news headlines \cite{Roumeliotis2024LLMsCrypto}. Such models can discern subtle tone in news (e.g. regulatory announcements or macroeconomic signals) and predict short-term market reactions. In practical terms, positive or negative sentiment scores produced by LLMs serve as features for trading algorithms or risk management, often improving predictive power over using price data alone. Beyond news, researchers are leveraging LLMs to \textbf{analyze social media} (tweets, Reddit discussions) for crypto sentiment, which is a multimodal challenge: understanding slang, memes, and the context of discussions requires advanced language comprehension.

LLMs are also being explored as direct \textbf{time-series forecasters} for crypto prices. In a novel approach, Makri \textit{et al.}\ (2025) repurposed a pre-trained language model for Ethereum price prediction by treating historical price sequences as the input ``language''~\cite{makri2025ethereumpricepredictionemploying}. By freezing most of the language model's layers and fine-tuning it on a limited sequence of price data, they achieved state-of-the-art accuracy in short-term forecasting (outperforming classic statistical models on MSE and RMSE). This suggests LLM architectures can capture complex temporal patterns, possibly learning latent market dynamics.
Recent research on fact-subjectivity--aware reasoning agents shows that GPT-4--powered analytics can track sentiment shifts across multiple social platforms, providing real-time alerts to fear-or-hype cycles~\cite{wang2025exploringllmcryptocurrencytrading}.
Moving forward, researchers suggest \textbf{multimodal fusion} of textual sentiment with numerical price data to further improve crypto forecasts. An LLM could, for example, read news headlines and concurrently analyze price charts, combining qualitative and quantitative signals---an approach aligned with how human analysts operate. The trend in this subdomain is toward \textbf{integrating diverse data modalities} (news, social chatter, on-chain metrics, price history) using LLMs as the unifying analytical engine. Key challenges include the need for up-to-date information (crypto news cycles are rapid), handling the noise and intentional manipulation in social media data, and ensuring that models do not learn spurious correlations. Nevertheless, early benchmarks show that domain-tuned LLMs can significantly enhance crypto sentiment analysis and market prediction accuracy~\cite{Roumeliotis2024LLMsCrypto}.

\subsubsection{Fraud Detection and Anomaly Detection}
Blockchain's open ledger provides a rich data source for analyzing transactions, but spotting illicit or fraudulent activities in massive transaction streams is a needle-in-haystack problem. LLMs have emerged as powerful \textbf{anomaly detectors} in this context by learning the normal patterns of blockchain transactions and flagging outliers. Unlike rule-based fraud systems (which rely on known signatures of scams), an LLM can generalize from contextual cues and detect \textbf{previously unseen attack patterns} \cite{he2025largelanguagemodelsblockchain}. For example, \textit{BLOCKGPT} is a Transformer-based model trained on millions of Ethereum transactions (with a special focus on known attack cases). It operates as an IDS (Intrusion Detection System) for blockchain, monitoring incoming transfers and smart contract calls in real-time. Impressively, BLOCKGPT was able to correctly identify 49 out of 124 real exploit transactions among the top anomalies in testing, catching complex attacks that would evade simpler detectors. Moreover, it maintained high throughput – analyzing $\sim$2,284 transactions per second – indicating viability for deployment in live networks. The success of such models comes from a \textbf{proactive, learning-based approach}: the LLM encodes a broad notion of ``normal'' vs ``suspicious'' behavior rather than any single indicator. Other efforts use LLMs to examine blockchain data for money laundering patterns, Ponzi schemes, or phishing attempts by interpreting transaction histories as narratives. Multimodal extensions are also possible here (e.g. combining on-chain data with darknet forum text where criminals coordinate).

A related application is \textbf{log analysis in distributed networks}, where LLMs parse system logs or protocol messages to detect anomalies or predict failures. For instance, LogGPT and similar models have been applied to general IT system logs, demonstrating that an LLM can outperform traditional anomaly detectors by understanding semantic context in events \cite{guan2025logllmlogbasedanomalydetection,sinha2024realtimeanomalydetectionreactive}. In blockchain, this means beyond financial fraud, LLMs can help spot network malfunctions or security breaches by reading the sequence of node logs or consensus messages. The trend in anomaly detection is leaning toward \textit{real-time LLM-driven monitors} that continuously learn from new data (staying adaptive to new hacking strategies). Challenges include scalability (running large models continuously can be costly, though lighter fine-tuned models or distillations are being studied) and the availability of labeled anomalies for training. There is also a risk of adversarial attempts to fool LLM-based detectors, prompting research into robustifying these models for security. Nonetheless, early deployments like BLOCKGPT highlight that LLMs bring a \textbf{significant advantage in flexibility and depth of understanding} for blockchain fraud detection, potentially making decentralized ecosystems safer.

\subsubsection{Decentralized Governance and Decision-Making}
Decentralized communities (e.g. DAO governance forums, crypto social platforms) generate large volumes of unstructured text – proposals, discussions, votes – which LLMs are well-suited to analyze and support. One emerging area is using LLMs as \textbf{participants or assistants in blockchain governance}. Research by Trozze \textit{et al.} explored GPT-3.5's ability to assist in legal reasoning for crypto-related cases, like identifying regulatory violations in a token sale scenario \cite{trozze2024largelanguagemodelscryptocurrency}. The LLM could correctly suggest some applicable laws, though it missed others, highlighting that current models still struggle with complex legal logic. Interestingly, the same work showed ChatGPT can draft legal documents (e.g. a lawsuit complaint) of comparable quality to human lawyers – mock jurors could not distinguish AI-written complaints from human-written ones. This suggests LLMs could aid in writing proposals, bylaws, or explaining governance policies in crypto communities, improving efficiency. Similarly, \textbf{community moderation} is being augmented by LLMs: Axelsen \textit{et al.} (2023) demonstrate that ChatGPT-based systems can enforce forum rules and filter toxic content in online communities \cite{he2025largelanguagemodelsblockchain}. In a DAO context, this means automating the removal of spam or off-topic posts and summarizing member comments, ensuring healthier deliberation. Another study proposes an \textbf{LLM-driven blockchain governance framework}, where an LLM helps classify and route proposals, and even provides recommendations by drawing on past proposals and outcomes. These AI ``advisors'' in governance could democratize knowledge – e.g. a DAO member could query an LLM about the implications of a complex tokenomics change and get an accessible summary rather than reading a 50-page whitepaper.

While promising, \textbf{decentralized decision-making support} via LLMs faces challenges. One issue is \textbf{domain understanding} – models must grasp technical and economic context (for example, the nuances of a liquidity pool's tokenomics) to give sound advice. Misunderstandings could lead to flawed recommendations. There are also concerns about trust and autonomy: current research cautions that LLM assistants should remain advisory, not autonomous decision-makers, until they prove reliability \cite{he2025largelanguagemodelsblockchain}. Bias is another worry; an LLM trained predominantly on English crypto discussions might undervalue perspectives from other language communities, affecting truly global projects. Privacy and confidentiality matter as well, especially for legal and financial advice – models need to handle sensitive governance data without leaking. Despite these hurdles, the trend is that \textbf{LLMs are increasingly integrated into the workflows of crypto communities} – from drafting governance proposals and smart contracts, to providing real-time legal/compliance feedback, to guiding newcomers through educational tools (e.g. \textit{GPTutor} for blockchain finance FAQs). In sum, LLMs act as amplifiers of human expertise in decentralized settings, but careful oversight is required. As models improve in factual accuracy and context-awareness, we can expect them to take on larger supportive roles in decentralized decision-making, potentially enabling more informed and inclusive governance.

\subsubsection{Section Conclusion}
\textbf{General trends:} Across blockchain and cryptocurrency applications, LLMs are proving to be versatile allies – excelling at synthesizing code and language, which is crucial in a domain that blends software with finance and law. We observe a surge of domain-specific LLM efforts in 2023–2025 tackling everything from smart contract security to market prediction. A common theme is integration: rather than using LLMs in isolation, the best results come from combining them with other tools (e.g. static analyzers in security, or financial models in trading) and modalities (text, code, numeric data). Multimodal LLMs, which can handle not just text but also sources like transaction graphs or time-series, are on the horizon to address the rich data in crypto ecosystems. 

\textbf{Challenges:} Despite progress, significant challenges remain. Ensuring \textbf{accuracy and reliability} is paramount – a single hallucinated output in a contract audit or an erroneous trading signal can cause large losses. This drives the need for benchmark datasets and rigorous evaluation: as noted, many studies currently use bespoke datasets (Table~\ref{tbl:finance_discipline}), making it hard to compare models. Creating standardized, community-accepted \textbf{benchmarks for LLM4Crypto} tasks (for vulnerability detection, fraud detection, etc.) will be important for measuring progress. Scalability is another concern: blockchain data is continuously growing, and LLM inference can be costly; efficient fine-tuning and model compression will be needed for practical deployment. \textbf{Privacy and ethical considerations} also come to the forefront. In decentralized finance, an LLM might inadvertently reveal sensitive information from training data, or contribute to market manipulation if misused for fake news generation – calling for careful policy and perhaps on-chain verification of AI outputs.

Nonetheless, the \textbf{future outlook} is optimistic. LLMs are rapidly improving their capacity to reason about structured data (like code and transactions) and can be expected to become even more accurate with domain adaptation. We anticipate LLMs playing a key role in \textbf{automated blockchain development lifecycles} – for example, an AI assistant that can write a smart contract, formally verify it, deploy it, and monitor its transactions, all through natural language instructions. In crypto finance, LLMs might evolve into personalized financial advisors that understand a user's portfolio and the market sentiment simultaneously. The interdisciplinary nature of this domain means collaboration between AI researchers, security experts, economists, and the crypto community is vital. In conclusion, LLMs are catalyzing a new wave of intelligence in blockchain technology, \textbf{enhancing security, transparency, and decision-making}. As research converges on better benchmarks and methods, we move closer to realizing trustworthy AI-driven systems in decentralized finance – a step that could bolster confidence and accelerate innovation in the blockchain space.


\section{Mathematics}\label{sec:mathematics}
Mathematics benchmarks evaluate LLMs' capabilities in formal reasoning, theorem proving, and quantitative analysis across various mathematical domains. These benchmarks test both computational accuracy and the depth of conceptual understanding. An overview of all the relevant benchmarks is presented in Table~\ref{tbl:mathematics_discipline}.

\begin{table*}[!ht]
  \caption{Mathematics Discipline: Domain-Specific Benchmarks}\label{tbl:mathematics_discipline}
  \centering
  
  \resizebox{\fulllength}{!}{%
  \begin{tabular}{lllcccccc} 
  \cmidrule[\heavyrulewidth]{1-9}
  \textbf{Domain} & \textbf{Sub-domain} & \textbf{Benchmark} & \textbf{Scale} & \textbf{Task Type} & \textbf{Input Modality} & \textbf{Model} & \textbf{Performance} & \textbf{Key Focus} \\
  \cmidrule[\heavyrulewidth]{1-9}
  \multirow{6}{*}{Mathematical Reasoning} 
  & \multirow{6}{*}{Problem Solving}
  & MathVerse~\cite{zhang2024good} & \textcolor{quantified}{15K samples} & Visual math & Image + Text & GPT-4V & \textcolor{unspecified}{N/A} & Diagram interpretation \\
  && HARDMath~\cite{fan2024hardmath} & \textcolor{quantified}{366 problems} & Graduate-level & Text & GPT-4 & 43.8\% & Advanced reasoning \\
  && MMIQC~\cite{liu_augmenting_2024} & \textcolor{unspecified}{N/A} & Competition math & Text & \textit{Qwen-72B-MMIQC} & 45.0\% & Question composition \\
  && MathChat~\cite{liang2024mathchat} & \textcolor{unspecified}{N/A} & Interactive & Text + Dialogue & \textcolor{unspecified}{N/A} & \textcolor{unspecified}{N/A} & Multi-turn reasoning \\
  && GSM-MC~\cite{zhang10multiple} & \textcolor{unspecified}{N/A} & Multiple-choice & Text & Multiple & 30x faster & Efficient evaluation \\
  && PROBLEMATHIC~\cite{anantheswaran2024cutting} & \textcolor{unspecified}{N/A} & Robustness & Text & Llama-2 & +8\% & Noise handling \\
  \cmidrule[\heavyrulewidth]{1-9}
  \multirow{1}{*}{Formal Methods} 
  & \multirow{1}{*}{Theorem Proving}
  & LeanDojo~\cite{yang2023leandojo} & \textcolor{quantified}{98,734 theorems} & Theorem proving & Text (code) & GPT-4 & \textcolor{unspecified}{N/A} & Formal verification \\
  \cmidrule[\heavyrulewidth]{1-9}
  \multirow{1}{*}{Statistical Analysis} 
  & \multirow{1}{*}{Data Analysis}
  & KnowledgeMath~\cite{zhao2024financemath} & \textcolor{quantified}{1,259 problems} & Finance MWPs & Text + Tables & GPT-4 & 45.4\% & Domain knowledge \\
  \cmidrule[\heavyrulewidth]{1-9}
  \end{tabular}%
  }
  
  \end{table*}

\subsection{Mathematical Reasoning -- Problem Solving}
MATHVERSE is a benchmark specifically designed to evaluate the mathematical reasoning capabilities of language models in situations that involve interpreting complex visual content, such as diagrams and geometric figures~\cite{zhang2024good}. Unlike the traditional benchmarks that rely only on text-based prompts, MATHVERSE focuses on multimodal reasoning, where both textual descriptions and visual elements must be synthesized to solve complex problems. This kind of reasoning mirrors the real-world mathematical problem-solving skills exhibited by learners, where visuals often play a crucial role. The benchmark's structure highlights a significant gap in current large language model design, where many of these models underperform on visual-based tasks, indicating a lack of robust visual understanding in even the most advanced systems.

To support a deeper analysis of how models handle visual versus textual reasoning, MATHVERSE introduces six controlled variations of each problem by iteratively modifying or removing textual and diagrammatic elements. This design enables researchers to isolate the contributions of each modality to the model's final answer. Interestingly, the evaluation reveals that some LLMs achieve higher accuracy when diagrams are excluded, suggesting that these models may be relying on textual cues and redundancies rather than truly interpreting the visual information. This observation raises concerns about the depth of multimodal reasoning in current models and emphasizes the need for improved integration of visual understanding within language models to handle authentic, diagram-rich mathematical tasks effectively~\cite{zhang2024good}.

HARDMath~\cite{fan2024hardmath} represents a significant step forward in evaluating the mathematical reasoning capabilities of large language models by shifting the focus from elementary problem-solving to rigorous, research-grade challenges. Unlike earlier datasets that predominantly featured deterministic computations or straightforward symbolic manipulation, HARDMath introduces tasks that demand a deeper conceptual understanding and the ability to navigate open-ended problem spaces. This shift is critical in assessing a model's aptitude for graduate-level reasoning, where precision, abstraction, and strategic thinking are essential. By incorporating complex domains such as asymptotic behavior and approximation theory—areas where closed-form solutions are often unavailable—HARDMath pushes models to emulate the thought processes of human mathematicians rather than simply following formulaic procedures.
HARDMath's~\cite{fan2024hardmath} design not only increases the difficulty of tasks but also enhances the transparency and reproducibility of model evaluation. The inclusion of automatically generated, step-by-step solutions allows researchers to track the reasoning paths models take, making it easier to diagnose errors and understand performance bottlenecks. This structured approach to evaluation provides a more granular insight into a model's strengths and weaknesses, facilitating targeted improvements. As a result, HARDMath not only raises the bar for what it means to ``understand'' mathematics in the context of AI but also offers a robust framework for future developments in mathematically capable language models.

MMIQC~\cite{liu_augmenting_2024} introduces a high-quality dataset for competition-style mathematics, generated through the novel Iterative Question Composing (IQC) method, which refines math problems in stages and employs LLM-based rejection sampling to ensure clarity, difficulty, and pedagogical value. This iterative augmentation process produces problems that closely resemble real math competition questions, offering both diversity and depth. Fine-tuning large open-source models such as Qwen-72B on MMIQC results in substantial performance improvements, with reported gains of over 8\% on the MATH benchmark—highlighting the effectiveness of domain-specific data in enhancing advanced reasoning abilities and setting a new standard for dataset generation in mathematical problem solving.

The MathChat benchmark~\cite{liang2024mathchat} offers a new measure by which to assess the depth and adaptability of LLMs when confronted with mathematical reasoning tasks that unfold over multiple conversational turns. These types of problems both test a model's computational accuracy and its ability to manage context, interpret user intent over time, and generate intermediate steps that guide the reasoning process. Unlike the traditional benchmarks, which focus on one-shot question-answering, MathChat emphasizes the need for LLMs to engage in dynamic, multi-step dialogue—mirroring how real users might approach complex math queries in practice. This includes the model's capacity to ask clarifying questions, revise earlier reasoning when new information emerges, and maintain consistency across a series of interdependent prompts.

In their report on the MathChat benchmark, Liang et al.~\cite{liang2024mathchat} present a comprehensive analysis of how various LLMs handle these challenges, revealing a substantial performance gap among them. While GPT-4o demonstrated strong contextual retention and error-correction capabilities, particularly in maintaining logical consistency over multiple exchanges, other models struggled with either misinterpreting follow-up queries or failing to revise incorrect prior responses. These findings shine a light on the importance of interactive reasoning benchmarks like MathChat in advancing the development of more robust and conversationally intelligent LLMs, especially in domains such as mathematics where accuracy, adaptability, and sustained reasoning are critical.

The Grade School Math - Multiple Choice (GSM-MC) benchmark measures both the accuracy and speed of LLMs when solving grade school-level math problems presented in a multiple-choice format. Proposed by Zhang et al.~\cite{zhang10multiple} as an enhancement of the GSM8K dataset, GSM-MC introduces a structured question format that not only tests computational correctness but also the models' ability to discern between correct answers and plausible distractors—a key challenge in educational and standardized testing contexts. Unlike traditional benchmarks that focus solely on accuracy, GSM-MC incorporates timing metrics to assess computational efficiency, making it particularly relevant for applications where rapid inference is essential. Additionally, this benchmark serves as a diagnostic tool for identifying and addressing known issues in LLM behavior, such as biases toward certain response options, a problem previously noted by Pezeshkpour and Hruschka~\cite{pezeshkpour-hruschka-2024-large}. By utilizing randomized and well-balanced multiple-choice sets, GSM-MC provides a more nuanced understanding of a model's reasoning capabilities, helping distinguish genuine problem-solving ability from reliance on statistical patterns, and offering valuable guidance for future improvements in model training and evaluation methodologies.

PROBLEMATHIC~\cite{anantheswaran2024cutting} is a benchmark evaluation dataset developed by researchers at Arizona State University and Georgia Institute of Technology to examine the robustness of LLMs when solving real-world math word problems, including both simple problems that require a single operation as well as more complex, multi-step problems. Unlike the traditional mathematical reasoning benchmarks, such as MathChat, which primarily test a model's ability to compute correct answers, PROBLEMATHIC is unique in emphasizing the importance of parsing through irrelevant or misleading information - often presented in adversarial prompts - to assess whether models can identify and focus on the facts necessary to solve a problem. This design aligns with the cognitive demands placed on human problem solvers and provides a more realistic evaluation of an LLM's reasoning abilities. The benchmark reveals a significant vulnerability in current models, with performance dropping by up to 26\% when adversarial noise is introduced, underscoring the difficulty LLMs face in distinguishing between relevant and irrelevant information. However, the dataset also highlights the potential of targeted training approaches: fine-tuning on adversarially noisy examples improves model robustness by approximately 8\%, illustrating the value of noise-aware training strategies and the need for more sophisticated reasoning mechanisms in LLM development.

\subsection{Formal Methods -- Theorem Proving}

LeanDojo~\cite{yang2023leandojo} serves as a comprehensive and challenging benchmark for testing the capabilities of large language models in the domain of mathematical theorem proving. By providing 98,734 theorems and proofs extracted from a vast collection of 3384 Lean proof assistant files, it allows developers to assess how well their models perform on tasks related to formal mathematics. The Lean proof assistant itself is an open-source tool widely used by researchers to formalize and verify mathematical proofs. The open availability of LeanDojo as a Lean playground—where all the necessary data, resources, and evaluation criteria are included—ensures that researchers and developers have a consistent and transparent framework to compare the performance of different LLMs in proving theorems. One of the key insights from the benchmark's initial results is that many current LLMs struggle significantly with mathematical reasoning tasks, as seen in the relatively low success rate of models like ReProver, which only correctly proved 51.2\% of the theorems.

The LeanDojo benchmark's structure has made it a valuable tool for advancing the field of automated theorem proving. Its ability to enable models to interact directly with the Lean proof assistant allows for a more nuanced evaluation of model performance, particularly in stepwise proof construction. Additionally, LeanDojo's retrieval-augmented model, ReProver, highlights the importance of premise selection—demonstrating that effective identification of relevant premises can substantially boost a model's theorem proving accuracy. Furthermore, the benchmark's design includes a split that tests models on previously unseen theorems requiring novel premises, which ensures that the evaluation process measures the model's ability to generalize beyond what it has encountered during training. This approach not only promotes a deeper understanding of how well models can handle mathematical reasoning but also supports the broader goal of improving LLM performance in structured, formal problem-solving environments ~\cite{labelle2024monte,yang2023leandojo}.

\subsection{Statistical Analysis -- Data Analysis}

KnowledgeMath, also referred to in some publications as FinanceMath, was introduced by Zhao et al.~\cite{zhao2024financemath} as a comprehensive benchmark designed to assess the capabilities of large language models in solving complex mathematical challenges related to the financial sector. The benchmark includes over 1200 problems relating to a range of topics, including financial mathematics and statistical analysis, presented in diverse formats such as textual descriptions and graphical tables. Approximately 40\% of these tasks require tabular reasoning, demanding a high level of analytical thinking and problem-solving. To support more detailed insights into the models' reasoning abilities, each problem is paired with executable Python-based solutions, which enable code-level analysis and verification of the model' approach. This structure makes it possible to directly compare LLM performance against human experts, who achieve an average accuracy of 94\%. The benchmark thus provides a rigorous standard for evaluating model performance.

At the time that KnowledgeMath was released, even the most advanced models, such as GPT-4, were found to underperform significantly, achieving less than 50\% accuracy on the benchmark. This gap in performance quality demonstrated the limitations of even state-of-the-art models in handling specialized tasks requiring advanced mathematical reasoning and financial domain knowledge. However, recent developments have led to the creation of fine-tuned models that show promise in improving performance. For instance, models like GPT-4o and Gemini, when augmented with Chain-of-Thought and Program-of-Thought prompting techniques, have demonstrated considerable improvements, with GPT-4o reaching an accuracy of 60.9\%. KnowledgeMath's integration of a domain knowledge base encompassing over 860 finance-related concepts further enhances the LLMs' ability to retrieve relevant information, contributing to their improved performance. Yuan et al.~\cite{yuan2024finllms} highlight these advances, showcasing how specialized prompting strategies and domain knowledge integration are key to refining LLM capabilities in the financial sector.

\section{Humanities}\label{sec:humanities}
The Humanities discipline encompasses diverse fields that examine human culture, society, and creative expression. Recent benchmarks in this area evaluate LLMs' capabilities in understanding social dynamics, artistic creation, and ethical reasoning. An overview of all the relevant benchmarks is presented in Table~\ref{tbl:humanities_discipline}.

\begin{table*}[!ht] 
  \caption{Humanities Discipline: Domain-Specific Benchmarks}\label{tbl:humanities_discipline}
  \centering
  
  \resizebox{\fulllength}{!}{%
  \begin{tabular}{lllcccccc} 
  \cmidrule[\heavyrulewidth]{1-9}
  \textbf{Domain} & \textbf{Sub-domain} & \textbf{Benchmark} & \textbf{Scale} & \textbf{Task Type} & \textbf{Input Modality} & \textbf{Model} & \textbf{Performance} & \textbf{Key Focus} \\
  \cmidrule[\heavyrulewidth]{1-9}
  \multirow{9}{*}{Social Studies} 
  & \multirow{3}{*}{Social Media Analysis}
  & MM-SOC~\cite{jin2024mm} & \textcolor{collection}{Multiple} & Social media & Multi-modal & Multiple & \textcolor{unspecified}{N/A} & Platform analysis \\
  && HOTVCOM~\cite{chen_hotvcom_2024} & \textcolor{quantified}{10K videos} & Comment generation & Video + Text & Multiple & 0.42 ROUGE-L & User engagement \\
  && XMeCap~\cite{chen_xmecap_2024} & \textcolor{quantified}{10K memes} & Caption generation & Multi-image & Multiple & 0.31 BLEU & Meme creation \\
  \cmidrule{2-9}
  & \multirow{2}{*}{Bias and Privacy}
  & Priv-IQ~\cite{shahriar_priv-iq_2025} & \textcolor{collection}{Multiple} & Privacy intelligence & Multi-modal & Multiple & \textcolor{unspecified}{N/A} & Privacy evaluation \\
  && LLM-Bias~\cite{gallegos_bias_2024} & Survey paper & Bias analysis & Text & Multiple & \textcolor{unspecified}{N/A} & Fairness evaluation \\
  \cmidrule{2-9}
  & \multirow{2}{*}{Sociocultural}
  & CultureVLM~\cite{liu_culturevlm_2025} & \textcolor{quantified}{100K images} & Cultural understanding & Image + Text & Multiple & \textcolor{unspecified}{N/A} & Cultural diversity \\
  && TimeTravel~\cite{ghaboura_timetravel_2025} & \textcolor{collection}{Multiple} & Historical reasoning & Multi-modal & Multiple & \textcolor{unspecified}{N/A} & Cultural heritage \\
  \cmidrule{2-9}
  & \multirow{2}{*}{Emotional}
  & EmoBench-M~\cite{hu_emobench-m_2025} & \textcolor{collection}{Multiple} & Emotion cognition & Multi-modal & Multiple & \textcolor{unspecified}{N/A} & Emotional intelligence \\
  && EmotionQueen~\cite{chen_emotionqueen_2024} & \textcolor{quantified}{1,000 dialogues} & Empathy evaluation & Text & Multiple & \textcolor{unspecified}{N/A} & Empathetic response \\
  \cmidrule[\heavyrulewidth]{1-9}
  \multirow{3}{*}{Arts \& Creativity} 
  & \multirow{3}{*}{Creative Tasks}
  & EditWorld~\cite{yang_editworld_2024} & \textcolor{collection}{Multiple} & Image editing & Image + Text & Multiple & \textcolor{unspecified}{N/A} & World dynamics \\
  && LLM-Narrative~\cite{he_llms_2024} & Survey paper & Storytelling & Text & Multiple & \textcolor{unspecified}{N/A} & Narrative generation \\
  && WenMind~\cite{cao_wenmind_2025} & \textcolor{collection}{Multiple} & Creative expression & Multi-modal & Multiple & \textcolor{unspecified}{N/A} & Emotional creativity \\
  \cmidrule[\heavyrulewidth]{1-9}
  \multirow{5}{*}{Music} 
  & \multirow{5}{*}{Music Intelligence}
  & ZIQI-Eval~\cite{li_music_2024} & \textcolor{quantified}{14K items} & Music knowledge & Text & GPT-4 & 58.68\% F1 & Comprehensive music \\
  && MuChoMusic~\cite{weck_muchomusic_2024} & \textcolor{quantified}{1,187 questions} & Audio-language & Audio + Text & Multiple & \textcolor{unspecified}{N/A} & Music comprehension \\
  && Music-LLM~\cite{zhou_can_2024} & \textcolor{collection}{Multiple} & Symbolic music & Text & Multiple & \textcolor{unspecified}{N/A} & Music reasoning \\
  && MER-Benchmark~\cite{hachmeier_benchmark_2025} & \textcolor{collection}{Multiple} & Entity recognition & Text & Multiple & \textcolor{unspecified}{N/A} & Music entities \\
  && MuChin~\cite{wang_muchin_2024} & \textcolor{collection}{Multiple} & Music description & Text & Multiple & \textcolor{unspecified}{N/A} & Chinese music \\
  \cmidrule[\heavyrulewidth]{1-9}
  \multirow{6}{*}{Urban Planning} 
  & \multirow{3}{*}{City Environment}
  & CityEQA~\cite{zhao_cityeqa_2025} & \textcolor{collection}{Multiple} & Embodied QA & 3D + Text & Multiple & 60.7\% human & Urban navigation \\
  && TransGames~\cite{zhang_transportation_2024} & \textcolor{collection}{Multiple} & Transportation & Multi-modal & Multiple & \textcolor{unspecified}{N/A} & Traffic analysis \\
  && LLM-Transport~\cite{nie_exploring_2025} & Survey paper & Transportation & Multi-modal & Multiple & \textcolor{unspecified}{N/A} & System integration \\
  \cmidrule{2-9}
  & \multirow{3}{*}{Urban Intelligence}
  & Urban-FM~\cite{zhang_urban_2024} & Survey paper & Urban modeling & Multi-modal & Multiple & \textcolor{unspecified}{N/A} & Foundation models \\
  && UrbanPlanBench~\cite{zheng_urbanplanbench_2024} & \textcolor{collection}{Multiple} & Urban planning & Text & Multiple & \textcolor{unspecified}{N/A} & Planning knowledge \\
  && CityBench~\cite{feng_citybench_2024} & \textcolor{quantified}{13 cities} & Urban simulation & Multi-modal & Multiple & \textcolor{unspecified}{N/A} & City management \\
  \cmidrule[\heavyrulewidth]{1-9}
  \multirow{5}{*}{Morality \& Ethics} 
  & \multirow{5}{*}{Moral Reasoning}
  & MoralBench~\cite{ji_moralbench_2024} & \textcolor{collection}{Multiple} & Moral evaluation & Text & Multiple & \textcolor{unspecified}{N/A} & Ethical reasoning \\
  && M$^3$oralBench~\cite{yan_m3oralbench_2024} & \textcolor{collection}{Multiple} & Multimodal ethics & Multi-modal & Multiple & \textcolor{unspecified}{N/A} & Visual morality \\
  && Greatest-Good~\cite{marraffini_greatest_2024} & \textcolor{collection}{Multiple} & Utilitarian reasoning & Text & Multiple & \textcolor{unspecified}{N/A} & Moral dilemmas \\
  && Value-Alignment~\cite{yao_value_2025} & \textcolor{collection}{Multiple} & Value systems & Text & Multiple & \textcolor{unspecified}{N/A} & Alignment evaluation \\
  && Self-Awareness~\cite{li_i_2024} & \textcolor{collection}{Multiple} & Self-knowledge & Text & Multiple & \textcolor{unspecified}{N/A} & Model capabilities \\
  \cmidrule[\heavyrulewidth]{1-9}
  \multirow{2}{*}{Philosophy and Religion} 
  & \multirow{1}{*}{Computational philosophy}
  & InterIDEAS~\cite{yang_interideas_2024} & \textcolor{quantified}{45,000 pages} & Intertextual link discovery & Text & Custom pipeline & 85-91\% accuracy & NLP-philosophy bridges \\
  \cmidrule{2-9}
  & \multirow{1}{*}{Religion and Theology}
  & Religion \& Chatbots~\cite{trepczynski_religion_2023} & \textcolor{unspecified}{N/A} & Philosophical reasoning & Text & Multiple LLMs & \textcolor{unspecified}{N/A} & Theological reasoning \\
  \cmidrule[\heavyrulewidth]{1-9}
  \end{tabular}%
  }
  
  \end{table*}

\subsection{Social Studies}
Social Studies benchmarks assess LLMs' ability to analyze social phenomena, cultural contexts, and human interactions. These evaluations reveal both promising capabilities and significant limitations in understanding complex social dynamics.

\subsubsection{Social Media Analysis}

Given the multimodal nature of social media user activity, the MM-Soc benchmark was introduced to evaluate the performance of multimodal LLMs (MLLMs) in diverse social media tasks across domains \cite{jin2024mm}. Evaluation of various tasks like misinformation detection and image description show notable limitations of MLLMs, particularly in zero-shot conditions, which are on par with random guesses. New datasets explore how to generate ``hot'' or popular comments for user engagement, reflecting a shift toward more multimodal cues \cite{chen_hotvcom_2024}. XMeCap addresses meme creation where multiple images must collectively convey humor or narrative \cite{chen_xmecap_2024}. Rather than handling single-image prompts, the authors dissect sub-image relationships and fuse them into a coherent caption. They also note the challenges in maintaining comedic timing and semantic cohesion when diverse visuals are at play.

\subsubsection{Bias and Privacy}

On the ethical side, comprehensive surveys map persistent concerns such as bias, privacy, and intellectual property while shedding light on emerging dilemmas like truthfulness and hallucinations \cite{deng_deconstructing_2024}. From a privacy perspective, the ability of LLMs to comprehend or demonstrate privacy intelligence is vital both in terms of ethics and its ability to serve as a privacy-preserving tool. To this end, Priv-IQ was proposed as the first multimodal privacy evaluating benchmark evaluating LLMs across eight privacy competencies like named entity recognition, multilingual privacy understanding, and knowledge of privacy regulations \cite{shahriar_priv-iq_2025}. Evaluations across state-of-the-art LLMs show that despite promising results, LLMs must improve in various competencies, including awareness of context and multilingual proficiency. Gallegos et al. \cite{gallegos_bias_2024} conducted a comprehensive review of bias and fairness in LLMs by introducing three taxonomies of metrics, datasets, and techniques for evaluating and mitigating bias.

\subsubsection{Sociocultural Understanding}

Multi-hop multimodal claim verification (MMCV) has emerged to address the rising complexity of cross-referencing textual, tabular, and visual information. Due to the rise in AI-generated content, automated fact-checking and verification are necessary. To this end, a large-scale claim verification dataset was introduced that reveals the challenges faced by MLLMs \cite{wang_piecing_2025}. Claim verification requires robust reasoning capabilities and reducing the risk of superficial or one-shot model conclusions. Given the importance of peer review for scientific publication, frameworks to simulate peer reviews and understand decisions are needed. To this end, Jin et al. \cite{jin_agentreview_2024} introduced an LLM-based agent peer review simulation framework. The authors found that reviewers' bias plays a notable role in peer review decisions, further explained by social influence theory. Recent multi-object sentiment analysis research demonstrates the growing complexity of multimodal tasks. Benchmarks like MOSABench \cite{song_mosabench_2024} highlight the challenges LLMs face when sentiment signals spread across multiple objects in an image. This points to the need for more advanced multi-object detection and context modeling. In parallel, multicultural understanding via large-scale datasets (like CultureVLM \cite{liu_culturevlm_2025}) shows increasing efforts to extend LLMs and VLMs beyond Western-centric data. These initiatives are designed to handle region-specific nuances (both textual and visual) so that models can adapt to diverse cultural norms and artifacts. TimeTravel brings attention to the extensive breadth and depth of historical artifacts; the framework categorizes and comprehends cultural pieces from multiple civilizations \cite{ghaboura_timetravel_2025}. Its core ambition is to see if AI can go beyond superficial pattern recognition to engage in historical reasoning to contribute to the digital preservation of cultural heritage. A survey of cultural representation and inclusion in LLMs reveals gaps in notable areas like semantic domains and aboutness and the need for analyzing cultural misrepresentation and underrepresentation \cite{adilazuarda_culture_2024}.

\subsubsection{Emotional Understanding}

Studies on emotion cognition indicate that despite notable strides, LLMs struggle with emotional subtlety and interpretability; ongoing challenges persist in context-sensitive emotion generation and classification \cite{chen_emotion_2024}. EmoBench-M introduces a novel viewpoint on how multimodal models manage intricate emotional cues across video, audio, and text, moving beyond simple sentiment tasks by incorporating more realistic and contextual scenarios \cite{hu_emobench-m_2025}. EmotionQueen focuses on aspects of empathy in large language models, particularly targeting how well they identify user motivations and handle implicit emotional contexts in conversational settings \cite{chen_emotionqueen_2024}.

\subsection{Arts \& Creativity}

Recent advancements in MLLMs have reshaped creative arts by introducing tools for artists and creators. EditWorld enables diffusion models to simulate realistic world dynamics based on human-like instructions \cite{yang_editworld_2024}. The authors curated a dataset with multimodal scenarios and used pretrained models like GPT-3.5. This approach enhanced models' capability to perform complex and scenario-aware editing tasks, surpassing traditional methods constrained by simpler operations. Recent research has extended the creative potential of MLLMs into the literary domain. For instance, the development of language-model-guided narrative generation frameworks that maintain long-term coherence and thematic depth, enabling novel storytelling approaches and complex narrative structures \cite{he_llms_2024}. These approaches leverage deep semantic understanding and contextual continuity, pushing creative boundaries beyond conventional language generation. Integrating generative AI techniques in interactive storytelling further demonstrates progress in creative domains. Specifically, advanced generative models have been employed to dynamically generate content within interactive narratives to enhance immersion and personalization of storytelling experiences \cite{chakrabarty_art_2024}. These AI-driven methods enable real-time adaptation of narrative elements based on user interactions. The potential of generative models is demonstrated to facilitate creative expression and actively engage participants in co-creative processes. Moreover, researchers explored the intersection of creative AI and emotional intelligence, developing systems that effectively manage emotional expressions and reactions in generated artworks and narratives \cite{cao_wenmind_2025}.

\subsection{Music}

A growing body of work is dedicated to benchmarking and evaluating the musical intelligence of LLMs, revealing insights and limitations. Focusing on symbolic music tasks, Zhou et al. \cite{zhou_can_2024} analyzed multi-step music reasoning and highlight that while LLMs like GPT-4 and LLaMA2 can generate structured outputs, they fail to synthesize motifs and forms. This indicates a deficiency in compositional logic of LLMs. Hachmeier and Jäschke \cite{hachmeier_benchmark_2025} explored music entity recognition in user-generated content. The authors noted that LLMs with in-context learning outperform smaller models but remain sensitive to prior exposure; therefore, there is room for improvements in generalization and mitigating hallucination. ZIQI-Eval was introduced as a large-scale music benchmark spanning over 14,000 items across ten thematic categories, from theory and performance to world music traditions \cite{li_music_2024}. Despite its breadth, results show all evaluated LLMs perform poorly, with GPT-4 scoring a modest average F1 of 58.68. This work demonstrates the gap between language-based reasoning and musical comprehension. A Chinese benchmark for colloquial music description reveals a dual challenge: LLMs must not only understand professional structures but also resonate with public sentiment \cite{wang_muchin_2024}. MuChoMusic \cite{weck_muchomusic_2024} can be used to evaluate audio-language models across 1,187 questions spanning knowledge and reasoning. The findings expose models' over-reliance on text cues and highlight weaknesses in cultural, lyrical, and temporal musical interpretation.

\subsection{Urban Planning}
Urban planning benchmarks evaluate LLMs' ability to understand and reason about city environments and transportation systems. These assessments reveal strengths in static geospatial tasks but expose challenges in real-time decision-making and interactive planning.

\subsubsection{City Environment and Transportation}

Urban intelligence and transportation modeling have emerged for evaluating LLM across multimodal, dynamic, and planning-oriented tasks. To this end, Zhao et al. \cite{zhao_cityeqa_2025} proposed a novel benchmark for embodied question answering in urban environments. The CityEQA benchmark requires agents to interpret spatial landmarks, navigate complex 3D cityscapes, and answer open-ended questions using a hierarchical framework. Despite achieving 60.7\% of human-level accuracy, LLMs face challenges with long-horizon spatial reasoning and fine-grained urban perception. Meanwhile, TransportationGames is a comprehensive benchmark for assessing LLMs and MLLMs on transportation tasks such as traffic regulation comprehension, accident analysis, and emergency plan generation \cite{zhang_transportation_2024}. The findings reveal that while text-based LLMs perform moderately well on memorization and understanding, multimodal tasks like sign recognition or road occupation detection expose gaps in practical reasoning and visual integration. Nie et al. \cite{nie_exploring_2025} examined the role of LLMs in transportation systems through a conceptual framework and survey. The authors categorize LLM functionalities into four roles (information processors, knowledge encoders, component generators, and decision facilitators) across sensing, learning, modeling, and managing tasks. Applications of LLMs include traffic prediction, autonomous driving, and safety analytics, and LLMs can act as adaptive agents in bridging fragmented pipelines and enabling context-aware urban mobility.

\subsubsection{Urban Intelligence}

Recent works have explored the promise of LLMs in urban intelligence. Zhang et al. \cite{zhang_urban_2024} proposed Urban Foundation Models, which are pre-trained on multimodal urban data. The authors introduced a taxonomy for classifying tasks and highlighting potential in domains like transportation, public safety, and environmental monitoring. The survey emphasizes the unique integration, spatio-temporal reasoning, and privacy challenges faced in building generalized urban intelligence models. UrbanPlanBench is a domain-specific benchmark evaluating LLMs across three key perspectives of urban planning: principles, professional knowledge, and regulations \cite{zheng_urbanplanbench_2024}. Despite leveraging a large instruction-tuned dataset, results reveal a gap between model performance and human-certified urban planners, especially in regulation-related tasks. CityBench provides a simulation-based evaluation framework to test LLMs on multimodal urban tasks such as traffic control, mobility prediction, and street navigation \cite{feng_citybench_2024}. Across 13 cities and multiple LLMs, results show strengths in static geospatial tasks but expose challenges in real-time decision-making and interactive planning.

\subsection{Morality and Ethics}

Across moral reasoning benchmarks, LLMs can identify ethical nuances but differ significantly in depth and consistency (\cite{ji_moralbench_2024}, \cite{yan_m3oralbench_2024}, \cite{marraffini_greatest_2024}, \cite{cheng_beyond_2025}, \cite{bulla_large_2025}). While some models, particularly the larger ones, tend to align with human intuitions (rejecting harm or maximizing collective good), others exhibit inconsistent judgment or require heavy prompting to produce coherent moral stances. The multimodal expansion of moral tasks further exposes how vision-based reasoning lags behind textual understanding, as classifying moral foundations from images adds complexity that current systems struggle to handle reliably. Advanced LLMs often outperform translation-based methods on cross-lingual tasks, yet cultural subtleties and domain shifts (like political discourse) remain stumbling blocks. Many studies converge on a similar pattern: top-tier LLMs handle general moral classification but break down on domain-specific or more nuanced ethical questions. Meanwhile, value alignment emerges as a multi-dimensional problem; no single model excels equally across all moral dimensions or value systems \cite{yao_value_2025}. Self-awareness--related tasks indicate that even the best LLMs can misstate their abilities, underscoring the tension between impressive moral or social reasoning and weak self-knowledge \cite{li_i_2024}.

\subsection{Philosophy and Religion}

Within broader philosophical and religious contexts, large corpora of historical and theological texts present a formidable test for advanced reasoning. In tasks like identifying intertextual links across extensive philosophical writings, a well-tuned LLM can approach expert-level accuracy \cite{yang_interideas_2024}. However, the success depends on domain-specific customization, showing that general-purpose chatbots need targeted fine-tuning for sophisticated tasks. Religious and theological queries demand interpretive depth ranging from layered scriptural exegesis to abstract metaphysical deduction, and current chatbot performance is inconsistent. They exhibit enough creativity to outline arguments or interpret symbolic language but often require significant prodding to attempt complex logic (and may still fail to see ramifications) \cite{trepczynski_religion_2023}. This domain reveals how advanced tasks like bridging theology and ontology test LLMs' ability to go beyond surface-level summarization into contextually rich reasoning.

\subsection{Section Conclusion}

Across the creative and civic domains, LLMs show remarkable progress but still fall short of expert-level reasoning and multimodal fluency (\cite{he_llms_2024}, \cite{chakrabarty_art_2024}). From short-story composition to Chinese classical poetry, specialized benchmarks (\cite{cao_wenmind_2025}) highlight the challenge of capturing nuanced cultural or artistic elements. In music, despite LLMs struggle when deep, domain-specific reasoning is required, particularly for advanced symbolic analysis or tasks that rely on actual audio content (\cite{zhou_can_2024}, \cite{hachmeier_benchmark_2025}). These works show that many models are prone to rely heavily on textual features rather than truly parsing or understanding non-text modalities like audio or symbolic notation. Across urban planning benchmarks, a common thread is the difficulty of bridging advanced domain knowledge (especially regulatory constraints, real-world city contexts, and multi-step reasoning) with the more general language abilities of LLMs (\cite{zhao_cityeqa_2025}, \cite{zheng_urbanplanbench_2024}, \cite{feng_citybench_2024}). While many models perform decently on basic city-related questions or straightforward tasks, their performance deteriorates when confronted with intricate traffic control, urban policy, and fine-grained geographical details. Across the humanities, although LLMs have begun to engage with human-centered domains of culture, arts, and urban systems, their ability to navigate nuanced and multimodal problems remains constrained. Continued advancements in fine-tuning, multimodal integration, and domain adaptation will be essential to close the performance gap.

\section{Finance}\label{sec:finance}

The finance domain presents unique challenges and opportunities for LLMs, requiring models to process diverse data types while maintaining high accuracy in high-stakes environments. This section explores how LLMs are being applied and evaluated across financial forecasting, sentiment analysis, document processing, and regulatory compliance. An overview of all the relevant benchmarks is presented in Table~\ref{tbl:finance_discipline}.

\begin{table*}[!ht]
\caption{Finance Discipline: Domain-Specific Benchmarks}\label{tbl:finance_discipline}
\centering

  \resizebox{\fulllength}{!}{%
\begin{tabular}{lllcccccc} 
\cmidrule[\heavyrulewidth]{1-9}
\textbf{Domain} & \textbf{Sub-domain} & \textbf{Benchmark} & \textbf{Scale} & \textbf{Task Type} & \textbf{Input Modality} & \textbf{Model} & \textbf{Performance} & \textbf{Key Focus} \\
\midrule
\multirow{8}{*}{Finance} & \multirow{3}{*}{Forecasting} & FLUE \cite{shah2022when} & 5 tasks & Sentiment, NER, classification & Text & FLANG-BERT, FinBERT & +3--5\% F1 & Financial text processing \\
 &  & FinMA / PIXIU \cite{xie2023pixiu} & 9 datasets & Classification, QA & Text + Structured & FinMA & +10--37\% F1 & Finance-specific fine-tuning \\
 &  & FinanceMath \cite{zhao2024financemath} & 1,259 items & Math reasoning & Text + Tables & GPT-4 & 60.9\% accuracy & Financial numerical reasoning \\
 & QA (Numerical) & FinQA \cite{chen2021finqa} & 8,281 pairs & Multi-step QA & Text + Tables & RoBERTa + executor & ~65\% exec acc. & Semi-structured math QA \\
 & QA (Long-context) & DocFinQA \cite{reddy2024docfinqa} & ~123k tokens & Document-level QA & Text + Tables & GPT-4 & ~67\% accuracy & Long-context computation \\
 & QA (Conversational) & ConvFinQA\cite{chen2022convfinqa} & 3,892 dialogues & Conversational QA & Text + Tables & FinQANet & 68.9\% exec acc. & Dialogue reasoning \\
 & QA (Factual) & FinanceBench \cite{islam2023financebench} & 10,231 items & Retrieval QA & Text + Retrieval & GPT-4 + retriever & 19\% correct & Hallucination resistance \\
 & ESG/Climate & ClimateBERT \cite{webersinke2021climatebert} & ~2M paras & Text classification & Text & DistilROBERTa & 46--48\% entropy loss & ESG and greenwashing \\
 & Info Extraction & FinRED \cite{sharma2022finred} & 2,400 articles & Entity/relation extraction & Text & FinBERT, RoBERTa & ~87\% F1 & Structured knowledge extraction \\
 & Multilingual QA & Golden Touchstone \cite{wu2024golden} & 12 datasets & Multi-task NLP & Text (bilingual) & GPT-4, FinGPT & Strong overall & Robust multilingual benchmarking \\
\cmidrule[\heavyrulewidth]{1-9}
\end{tabular}%
}

\end{table*}

\subsection{Introduction}

The domain of finance is particularly critical for deploying LLMs due to the high stakes of financial decision-making and the complexity of domain data, particularly considering the interaction of the various variables, but also the fact that, from a bottom-up perspective, these variables aggregate individual human behaviour \cite{subrahmanyam2008behavioural}.  Financial applications demand accurate understanding of diverse modalities, ranging from textual news and reports to structured balance sheets, time-series market data, charts, and potentially extending even to audio (e.g. earnings call recordings).  Unlike general text domains, financial analysis often requires reasoning over both narrative disclosures and quantitative data \cite{reddy2024docfinqa}.  Success in this domain can inform trading strategies, risk modelling, economic policy, macroprudential regulation and behavioural analysis for investor profiling \cite{rubbaniy2024dynamic,li2023large,karadima2021economic,hodbod2020sectoral,danielsson2016model}, but errors and false prediction can carry costly consequences.  Early studies show that even advanced models like GPT-4 falter on straightforward financial questions \cite{oehler2024does,islam2023financebench}, despite the model's apparent success in the academic realm \cite{dowling2023chatgpt}.  This dual nature that comprises rich multi-modal information and high reliability requirements makes economics and finance an interesting application domain for multimodal LLMs.

\subsection{LLM Applications and Benchmarks in Finance}

Financial applications of LLMs span multiple areas including market forecasting, sentiment analysis, document processing, and regulatory compliance. The following sections examine how specialized benchmarks evaluate model performance across these critical financial tasks.

\subsubsection{Financial Forecasting and Trading}

Forecasting asset prices and market movements is a core application area for LLMs in the financial domain.  This space has traditionally been dominated by time-series econometrics and, somewhat more recently, machine learning models, but is currently seeing growing interest in the integration of unstructured textual data with structured numerical inputs to capture market-relevant signals \cite{polyzos2024efficient}.  Multimodal LLMs offer new capabilities in this regard, enabling models to combine textual sources, such as financial news, earnings calls and analyst reports, with tabular or time-series data for predictive purposes \cite{kalamara2022making}  This fusion of multimodal input aligns with the behavioural finance literature, which recognises that investor decision-making is shaped not only by quantitative indicators but also by narratives and sentiment embedded in public communications \cite{polyzos2024efficient,li2023large,herrera2022renewable,dicks2021uncertainty}.  

Recent approaches solve the problem of multiple data sources by using multi-modal pipelines.  The Ploutos approach of Tong et al. \cite{tong2024ploutos} introduces a financial LLM framework with separate ``experts'' for text and numerical time-series, which are then combined to generate an interpretable trading rationale, which can support analysts' decisions.  Their results show that such hybrid LLM systems can outperform prior state-of-the-art trading algorithms on both prediction accuracy and explanation quality.  Similarly, open-source initiatives like FinGPT \cite{liu2023fingpt} and proprietary models like BloombergGPT \cite{wu2023bloomberggpt} have demonstrated data-centric techniques for algorithmic trading and robo-advising by automatically curating large volumes of financial data and fine-tuning models.  Finally, smaller versions, like FinMA and its accompanying benchmark suite PIXIU \cite{xie2023pixiu}, further contribute to this space.  This model has been trained on a curated corpus of Chinese and English financial data and has shown strong performance in financial sentiment analysis, document classification and QA tasks, validating the importance of domain alignment in financial forecasting applications.  On the computational side, an emerging benchmark is FinanceMath \cite{zhao2024financemath}, which evaluates LLMs on domain-specific math word problems rooted in financial contexts.

All these models leverage vast financial news archives, price data and even central bank statements.  Multimodality has recently extended to audio, with the newly introduced audio-text model, AT-FinGPT, that can fuse audio from earnings calls with transcript summaries for risk forecasting \cite{liu2025atfingpt}.  By capturing not only what executives say but how they say it (tone, sentiment), such models aim to predict volatility or default risk more accurately.  The model's early results in financial risk prediction confirm that integrating textual analysis with audio features calls can improve predictive power over text alone. 

Nonetheless, there are persistent limitations, with interpretability being the most prominent among them.  Golden Touchstone \cite{wu2024golden}, a new bilingual financial LLM benchmark, explicitly evaluates not only accuracy but also a model's ability to process complex information and provide useful explanations across tasks.  Temporal robustness also remains a challenge, as models trained on historical language patterns may not generalise well to sudden macroeconomic shifts or policy regime changes.  This is particularly relevant for markets affected by crisis events, even though unstructured data can often support capturing these crises \cite{polyzos2023inflation}.  Moreover, hallucination, outdated training data and, more importantly, misalignment with economic theory can lead to faulty or overconfident predictions, making it essential to pair LLMs with expert-guided guardrails.

\subsubsection{Sentiment Analysis and Market Signals}

Another active area for LLMs is sentiment analysis, where LLMs gauge the mood of investors or the tone of communications to inform investment decisions.  Financial markets are highly sensitive to sentiment in news, social media (e.g. Reddit, Twitter) and analyst reports \cite{long2023i,lyocsa2022yolo,polyzos2022quantifying,vasileiou2022does}.  Domain-specific models like FinBERT (a BERT variant trained on finance text) were early successes in parsing financial sentiment \cite{huang2023finbert}, building on the work of Loughran and McDonald \cite{loughran2011when}.  More recently, comprehensive benchmarks such as FLUE (Financial Language Understanding Evaluation) have catalysed progress \cite{shah2022when}.  FLUE is able to handle multiple tasks, including financial phrase sentiment classification, headline categorisation and Named Entity Recognition (NER), thus providing a suite of evaluation data to fine-tune and test LLMs on finance-specific language.  Models fine-tuned on these datasets can discern subtle sentiment cues, like differentiating optimistic vs. cautious tone in an earnings press release. For example, on the FiQA sentiment analysis task, a finance-tuned model (FLANG-BERT) significantly outperforms generic BERT, highlighting the value of domain adaptation \cite{shah2022when}. 

Such sentiment-aware LLMs have been applied to investor social media as well to quantify bullish or bearish signals.  In algorithmic trading, these sentiment signals are often combined with price data in a form of multimodal fusion where qualitative sentiment augments quantitative models.  There is also growing interest in behavioural finance modelling using LLMs, an example of which is analysing the narrative sentiment of central bank communications to predict market reactions \cite{nitoi2023unveiling}.  This bridges into behavioural finance, where LLMs help model how biases and emotions in text might foreshadow irrational market moves.  Sentiment analysis in finance demonstrates a success story for LLMs, since by training on domain-specific corpora and benchmarks, LLMs can achieve human-like interpretation of financial tone and emotion, providing valuable features for trading and risk assessment.

\subsubsection{Document Question Answering and Analysis}

Financial professionals frequently need to query large corpora of documents to extract specific information or perform due diligence.  LLMs have been applied to document question answering in this context, permitting integration of text with tabular data from financial statements.  A series of benchmark datasets have driven progress here. FinQA \cite{chen2021finqa} introduced a series of expert-written QA pairs on financial reports, focusing on numerical reasoning with both textual and tabular evidence.  Each question in FinQA is accompanied by a small Python computation, ensuring that models not only find facts but also perform calculations for full explainability.  Models are evaluated on answer accuracy and their ability to generate the correct reasoning program. 

Building on this, DocFinQA \cite{reddy2024docfinqa} extended the task to full annual reports, massively increasing context length.  This long-context setting reflects realistic analyst scenarios, such as browsing an entire filing for answers, and represents an apposite test of an LLM's ability to perform retrieval and reasoning over hundreds of pages.  Not surprisingly, it challenges even state-of-the-art models, as experiments found that GPT-4's error rate roughly doubled when moving from the short snippets of FinQA to the full-document DocFinQA setting.  The need to handle both natural language and semi-structured data (tables, financial figures) in a long context makes this a tough multimodal problem. 

Another variant is ConvFinQA, which transforms FinQA into a conversational QA format \cite{chen2022convfinqa} and can evaluate an LLM's ability to perform step-by-step reasoning in an interactive setting, thus mimicking an analyst probing a report iteratively.  Beyond academic datasets, industry has also recognised the need for robust QA evaluation with FinanceBench \cite{islam2023financebench}.  FinanceBench questions are designed to be answerable from filings with minimal ambiguity, thus providing essentially a minimum competency test for LLMs in finance.  Yet, when state-of-the-art models (GPT-4, LLaMA-2, etc.) were evaluated with retrieval augmented techniques, most struggled, which GPT-4 providing correct responses to only 19\% of questions and often refusing or hallucinating answers \cite{islam2023financebench}.  These QA benchmarks (the FinQA family and FinanceBench) have driven the progress but also emphasises the challenges.  LLMs fine-tuned on financial documents can answer many questions and even perform calculations, but ensuring accuracy and reliability at scale (especially for compliance-critical scenarios) demands increased model power but also consistency and precision.

\subsubsection{Regulatory Compliance, ESG and Behavioural Finance}

Beyond core trading and analysis tasks, LLMs are being explored in less mainstream but equally important subdomains of finance, with one key being regulatory compliance.  To ensure that they meet the required standards, institutions must parse complex regulations, policies and legal filings.  LLMs can assist by answering questions about compliance requirements or by flagging sections of a legal document relevant to a query.  While there is still a lack of large public benchmarks in this exact area, the capabilities overlap with the financial QA already discussed as well as with and legal NLP tasks.  As with most topics in the Finance domain, ensuring explainability is crucial here, since compliance officers need to trust and verify the sources of the answer provided by the LLM.  As a results, there is a strong need for LLMs that can cite evidence and reason transparently 

Another growing field is ESG (Environmental, Social, Governance) and climate finance. Specialised models like ClimateBERT \cite{webersinke2021climatebert} have been developed to handle climate-related financial text. ClimateBERT is a transformer model refined on sustainability reports and climate disclosures, enabling it to detect language about climate risks and commitments \cite{schimanski2024bridging}.  Tasks in this area include classifying statements in annual reports as substantive or ``greenwashing'', extracting metrics related to carbon emissions or summarising climate risk disclosures for investors.  Multimodal aspects appear when combining text with data like emissions figures or when analysing satellite imagery alongside reports for due diligence, though such extreme multimodality is still at early stages. 

Lastly, behavioural finance, which studies the psychological factors in investor decisions, offers intriguing opportunities for LLMs.  Researchers have begun using LLMs to model investor sentiment and biases or even to simulate market participants.  For example, an LLM agent could be assigned the persona of an ``overconfident investor'' versus a ``loss-averse investor'' to see how their responses differ when presented with the same news, potentially uncovering bias-driven divergences. While this is still exploratory, it builds on the idea of agent-based financial simulations that are able to uncover important insights regarding individual behaviour and how it can impact government policies \cite{polyzos2020who,polyzos2018good}.  A collaborative project by the Stevens Institute and others has started testing LLM agents in human-centred financial decision-making, essentially using LLMs to augment or simulate the role of financial advisors and investors \cite{technology2023applying}.  In this manner, domain-specific LLM research in this area aims to expand the scope of what such models can do, by moving from pure ``number-crunching'' and fact-finding into interpreting the human side of finance.

\subsection{Section Conclusion}

Across these subdomains, several clear trends have emerged.  First, we see a proliferation of finance-specific LLMs and fine-tuned models. Large proprietary models, like BloombergGPT and open-source counterparts, like FinGPT and FinMA/PIXIU, demonstrate the value of ingesting domain data at scale, continually integrating fresh financial data to keep models current.  Moreover, smaller models fine-tuned via instruction on financial tasks (e.g. FinMA on the FLUE benchmark tasks) have made domain adaptation accessible without training from scratch.  Second, there is growing emphasis on integrating multimodal capabilities with structured data handling, particularly in tasks that involve numerical reasoning.  In financial tasks that require multi-step calculations or querying large documents, such as those found in benchmarks like DocFinQA, models that combine LLMs with external tools tend to perform more accurately and reliably than those relying on language understanding alone.  Third, interpretability and explainability are recognised as essential features.  In high-stakes domains, such as Finance, a black-box answer is often not enough as regulators and practitioners demand reasoning traces. Approaches like program annotations in FinQA or rationale generation in Ploutos address this by design, and future benchmarks are likely to include explainability as a first-class metric.

Alongside these successes, however, significant challenges persist.  Access to high-quality financial data remains a significant challenge. Although the volume of financial information has increased substantially, much of it is proprietary or non-public, which restricts its use in open academic research \cite{liu2023fingpt}.  At the same time, publicly available data sources may contain misinformation, inconsistencies or unverified content, particularly when sourced from social media or news aggregators.  In addition, alignment with domain experts and industry regulations poses another hurdle, since an LLM's answers must not only be correct, but also compliant with legal and ethical standards.  Additionally, hallucination and factual accuracy are also important concerns.  This findings of FinanceBench regarding widespread mistakes even by top models underscore that current LLMs can be unreliable \cite{islam2023financebench}.  In applications like finance where factual errors can lead to financial loss or legal liability, this creates serious issues and hinders adoption.  Temporal robustness is also a unique challenge, as financial models need to stay current with evolving market conditions, new regulations and recent company disclosures, since an LLM trained on last year's data might miss critical context. Finally, efficiency and scalability are practical issues, since handling lengthy documents or streaming data in real-time can be computationally intensive and latency is a concern in trading environments.

Looking forward, several avenues could further advance domain-specific LLM performance in finance.  One is the creation of synthetic and privacy-preserving datasets.  For example, using simulators or generative techniques can help produce realistic financial data and scenarios that can augment training without risking real-world consequences. Another is developing standardised evaluation metrics for explainability and reasoning transparency, so that models are not only judged on accuracy but also on the faithfulness of their justifications.  Finally, integrating insights from economics and behavioural finance into model design could prove fruitful.  By embedding theoretical constraints or known bias patterns, we might guide LLMs to make decisions that are both rational and cognisant of human irrationality.  The trajectory in this domain suggests that finance will remain at the forefront of domain-specific LLM evaluation both because of its challenges and because of the tangible rewards of getting it right.

\section{Healthcare}\label{sec:healthcare}
Large Language Models (LLMs) have emerged as powerful tools in healthcare, with applications spanning clinical decision support, medical research, and patient care. Recent benchmarks demonstrate both the promise and limitations of these models across diverse medical domains. In clinical settings, LLMs show potential for assisting with diagnosis, treatment planning, and medical documentation~\cite{ye2024gmai,liu2024spectrum}. For bioinformatics applications, models are being evaluated on complex tasks like genomic analysis and drug discovery~\cite{keat2024pgxqa,liu2024genotex}. In medical imaging, specialized architectures are tackling challenges in radiology, pathology, and 3D analysis~\cite{bai_m3d_2024,bassi2025radgptconstructing3dimagetext}. An overview of all the relevant benchmarks is presented in Table~\ref{tbl:healthcare_discipline}.

\begin{table*}[!ht]
  \caption{Healthcare Discipline: Domain-Specific Benchmarks}\label{tbl:healthcare_discipline}
  \centering
  
  \resizebox{\fulllength}{!}{%
  \begin{tabular}{lllcccccc} 
  \cmidrule[\heavyrulewidth]{1-9}
  \textbf{Domain} & \textbf{Sub-domain} & \textbf{Benchmark} & \textbf{Scale} & \textbf{Task Type} & \textbf{Input Modality} & \textbf{Model} & \textbf{Performance} & \textbf{Key Focus} \\
  \cmidrule[\heavyrulewidth]{1-9}
  \multirow{9}{*}{Medicine \& Healthcare} 
  & \multirow{5}{*}{Clinical Applications}
  & GMAI-MMBench~\cite{ye2024gmai} & \textcolor{collection}{284 datasets} & VQA & Multi-modal & GPT-4o & 53.96\% & Comprehensive medical AI \\
  && Asclepius~\cite{liu2024spectrum} & \textcolor{quantified}{15 specialties} & Multi-task & Multi-modal & Multiple & \textcolor{unspecified}{N/A} & Specialty evaluation \\
  && MultiMed~\cite{mo_multimed_2024} & \textcolor{quantified}{2.56M samples} & Multi-task & Multi-modal & Multiple & \textcolor{unspecified}{N/A} & Cross-modality \\
  && MediConfusion~\cite{sepehri_mediconfusion_2024} & \textcolor{collection}{Paired images} & VQA & Multi-modal & Multiple & Below random & Model reliability \\
  && CTBench~\cite{neehal2024ctbench} & \textcolor{quantified}{2.5K samples} & Multi-task & Text & GPT-4 & 72.3\% & Clinical trial design \\
  \cmidrule{2-9}
  & \multirow{4}{*}{Bioinformatics}
  & PGxQA~\cite{keat2024pgxqa} & \textcolor{quantified}{110K+ questions} & QA & Text & GPT-4o & 68.4\% & Pharmacogenomics \\
  && GenoTEX~\cite{liu2024genotex} & \textcolor{collection}{Multiple datasets} & Gene analysis & Text + Data & GenoAgent & 65-85\% & Gene expression \\
  && MedCalc-Bench~\cite{khandekar2024medcalc} & \textcolor{quantified}{2K problems} & Calculation & Text + Numbers & GPT-4 & 57.8\% & Medical calculations \\
  && Bio-Benchmark~\cite{jiang2025benchmarking} & \textcolor{quantified}{30 tasks} & Multi-task & Text & GPT-4o & Varies by task & Bioinformatics NLP \\
  \cmidrule[\heavyrulewidth]{1-9}
  \multirow{8}{*}{Medical Imaging} 
  & \multirow{3}{*}{Radiology \& 3D Analysis}
  & M3D~\cite{bai_m3d_2024} & \textcolor{quantified}{120K img pairs} & Multi-task & 3D + Text & \textit{M3D-LaMed} & SOTA & 3D medical imaging \\
  && TriMedLM~\cite{10822809} & \textcolor{collection}{Multi-modal} & Multi-task & Image + Text & \textit{TriMedLM} & \textcolor{unspecified}{N/A} & Trimodal integration \\
  && RadGPT~\cite{bassi2025radgptconstructing3dimagetext} & \textcolor{quantified}{9.2K CT scans} & Report gen & 3D CT + Text & \textit{RadGPT} & 80-97\% sensitivity & Tumor detection \\
  \cmidrule{2-9}
  & \multirow{5}{*}{Pathology/Microscopy}
  & Micro-Bench~\cite{lozano2024micro} & \textcolor{quantified}{3.2K images} & VQA & Image + Text & GPT-4V & 54.7\% & Microscopy understanding \\
  && PathMMU~\cite{sun2024pathmmu} & \textcolor{quantified}{33K questions} & Multiple choice & Image + Text & GPT-4V & 49.8\% vs 71.8\% human & Pathology expertise \\
  && MicroVQA~\cite{burgess2025microvqamultimodalreasoningbenchmark} & \textcolor{quantified}{2.8K QA pairs} & VQA & Image + Text & GPT-4V & 48.2\% & Scientific microscopy \\
  && SlideChat~\cite{chen2025slidechatlargevisionlanguageassistant} & \textcolor{collection}{Large-scale} & Dialogue & Image + Text & \textit{SlideChat} & \textcolor{unspecified}{N/A} & Interactive analysis \\
  && $\mu$-Bench~\cite{lozano2024mubenchvisionlanguagebenchmarkmicroscopy} & \textcolor{collection}{Multiple tasks} & Multi-task & Image + Text & Multiple & Varies by task & Microscopy analysis \\
  \cmidrule[\heavyrulewidth]{1-9}
  \end{tabular}%
  }
  
  \end{table*}

\subsection{Medicine \& Healthcare}
The healthcare domain has seen extensive development of specialized benchmarks across clinical medicine, bioinformatics, and medical imaging. In clinical applications, benchmarks like GMAI-MMBench~\cite{ye2024gmai} and Asclepius~\cite{liu2024spectrum} evaluate models across diverse medical specialties and tasks. For bioinformatics, datasets like PGxQA~\cite{keat2024pgxqa} and GenoTEX~\cite{liu2024genotex} assess capabilities in genomics and pharmacology. Medical imaging introduces unique challenges, with benchmarks like M3D~\cite{bai_m3d_2024} and RadGPT~\cite{bassi2025radgptconstructing3dimagetext} testing 3D analysis capabilities. Let us examine each subdomain in detail.

\subsubsection{Clinical Applications}
Large Language Models (LLMs) are increasingly being applied in clinical healthcare, offering potential to assist medical professionals across diagnostic support, decision-making, and information access. Recent benchmarks have focused on evaluating how effectively multimodal LLMs can handle clinical tasks, revealing both promising capabilities and concerning limitations. These evaluations are critical for determining whether such models can safely support healthcare workflows, where errors can have serious consequences. The benchmarks in this domain test models across diverse medical specialties, task types, and input modalities, providing a comprehensive assessment of their clinical applicability and reliability.

GMAI-MMBench~\cite{ye2024gmai} represents the most comprehensive medical AI benchmark to date, comprising 284 datasets across 38 medical image modalities, 18 clinical tasks, and 18 medical departments. The benchmark features a well-categorized structure with four different perceptual granularities (image, box, mask, and contour levels), allowing for detailed assessment of model capabilities. Evaluations of 50 large vision-language models, including both open-source and proprietary systems, revealed that even the best performing model (GPT-4o) achieved only 53.96\% accuracy, highlighting significant room for improvement. The benchmark identified five key insufficiencies in current models: uneven task performance across different clinical scenarios, unbalanced department performance, inconsistent perceptual robustness, poor instruction following in medical contexts, and difficulties with multiple-choice questions.

Asclepius~\cite{liu2024spectrum} offers a specialized benchmark covering 15 medical specialties, designed to evaluate models' ability to perform across diverse clinical domains. This benchmark addresses the critical need for comprehensive evaluation across specialized medical fields, as performance in one specialty may not generalize to others. The evaluation identifies significant variations in model capabilities across different medical areas, highlighting the challenge of developing systems that maintain consistent performance across the spectrum of clinical practice.

MultiMed~\cite{mo_multimed_2024} provides a massive dataset of 2.56 million samples to assess models' capacity for cross-modality reasoning in medical contexts. This benchmark is particularly valuable for evaluating how effectively models can integrate information across different data types (such as images, text, and structured data), which is essential for many clinical workflows that require synthesis of diverse information sources. The large scale of this dataset enables robust assessment of model generalization capabilities across varied medical scenarios.

MediConfusion~\cite{sepehri_mediconfusion_2024} takes a unique approach by specifically probing the failure modes of medical multimodal LLMs. This benchmark exposes a concerning limitation: state-of-the-art models are easily confused by image pairs that are visually dissimilar and clearly distinct to medical experts. All evaluated models—both open-source and proprietary—performed below random guessing on this benchmark, raising serious questions about their reliability for clinical deployment. The study identified common patterns of model failure, including difficulties distinguishing between normal anatomy and pathology, interpreting lesion signal characteristics, analyzing vascular conditions, and identifying medical devices. These findings highlight fundamental limitations in current models' visual understanding capabilities.

CTBench~\cite{neehal2024ctbench} focuses specifically on clinical trial design, containing 2,500 samples to evaluate models' ability to assist in this critical aspect of medical research. GPT-4 achieved a relatively strong performance of 72.3\% on this benchmark, suggesting potential utility in supporting the complex process of clinical trial planning. This specialized benchmark addresses a specific clinical workflow where LLMs might provide significant value by accelerating research design and improving protocol quality.

\subsubsection{Bioinformatics}
Large Language Models offer transformative potential in bioinformatics by automating complex biological data analysis, interpreting genetic information, and accelerating biomedical research. Recent benchmarks in this domain evaluate LLMs' capabilities in processing specialized biological knowledge and performing data-intensive computational tasks. These evaluations reveal that while state-of-the-art models can achieve reasonable performance on specific bioinformatics tasks, there remains a substantial gap between current abilities and the expertise required for clinical implementation. The benchmarks in this area focus on assessing various capabilities essential to bioinformatics, from knowledge retrieval to data manipulation, providing insights into the strengths and limitations of applying LLMs to this highly specialized domain.

PGxQA~\cite{keat2024pgxqa} focuses on pharmacogenomics, evaluating LLMs on their ability to answer questions about genetics-guided treatment from the perspectives of clinicians, patients, and researchers. The benchmark contains over 110,000 questions covering areas such as allele definition, frequency, diplotype-to-phenotype mapping, and drug-gene interactions. Evaluation of models including GPT-4o, Llama 3, and Gemini 1.5 Pro revealed that while newer models like GPT-4o (68.4\% accuracy) significantly outperform older generations, they still fall short of the standards required for clinical use. The study highlights particular difficulties with questions related to allele definitions and functions, while showing stronger performance on queries involving drug and gene entities. This benchmark addresses a critical gap in pharmacogenetics implementation, where the lack of education and awareness among clinicians represents a major barrier to the adoption of precision medicine approaches.

GenoTEX~\cite{liu2024genotex} introduces a benchmark for evaluating LLM-based methods in automating gene expression data analysis. The benchmark provides analysis code and results for solving gene-trait association problems, covering key steps including dataset selection, preprocessing, and statistical analysis in a pipeline that adheres to computational genomics standards. The benchmark introduces three core tasks for evaluation: dataset selection, data preprocessing, and statistical analysis, each with corresponding metrics. To establish baselines, the researchers present GenoAgent, a team of collaborative LLM-based agents that adopt a multi-step programming workflow with flexible self-correction. Experimental results showed that GenoAgent performs well on certain tasks, achieving 65-85\% success across various metrics, demonstrating the potential of LLM-based methods in genomic data analysis. However, the evaluation also highlights limitations in handling complex tasks like clinical feature extraction and statistical analysis, identifying areas for further improvement.

MedCalc-Bench~\cite{khandekar2024medcalc} presents a benchmark consisting of 2,000 medical calculation problems designed to test LLMs' ability to perform precise quantitative analyses in healthcare contexts. The benchmark evaluates models on tasks requiring both medical knowledge and mathematical reasoning, with GPT-4 achieving a moderate performance of 57.8\%. These results highlight the challenge of combining domain expertise with computational accuracy, a critical requirement for clinical decision support. The benchmark addresses an important aspect of healthcare applications where errors in calculation could lead to serious consequences for patient care.

Bio-Benchmark~\cite{jiang2025benchmarking} offers a comprehensive framework for evaluating LLMs across 30 key bioinformatics tasks spanning seven domains: protein, RNA, RNA-binding protein (RBP), drug, electronic health records (EHR), medical Q\&A, and traditional Chinese medicine (TCM). The researchers evaluated six mainstream LLMs, including GPT-4o and Llama-3.1-70b, using both zero-shot and few-shot Chain-of-Thought settings without fine-tuning to reveal their intrinsic capabilities. The evaluation utilized BioFinder, a new tool developed for extracting answers from LLM responses with approximately 30\% greater accuracy than existing methods. Results revealed that few-shot learning significantly improved performance across multiple tasks, with some models seeing up to 20x improvement in protein species prediction and substantial gains in drug design and TCM tasks. Performance varied considerably across different bioinformatics subdomains, with GPT-4o generally performing best but different models excelling in specific areas—Mistral-large-2 led in protein species prediction (82\%) and drug design (91\%), while Llama-3.1-70b performed strongly in RNA function prediction (89\%).

\subsection{Medical Imaging}
Medical imaging benchmarks have emerged as a critical frontier in evaluating multimodal LLMs, particularly for their ability to handle complex 3D volumetric data and specialized imaging modalities. Recent evaluations reveal both promising advances and significant challenges - while models like RadGPT~\cite{bassi2025radgptconstructing3dimagetext} achieve strong performance in tumor detection (80-97\% sensitivity), even state-of-the-art systems struggle with basic visual reasoning tasks in pathology. The benchmarks in this domain are increasingly focusing on clinically-relevant capabilities, from 3D spatial understanding to interactive analysis workflows that mirror real clinical practice.

\subsubsection{Radiology \& 3D Analysis}
Medical imaging analysis presents unique challenges for multimodal large language models, particularly when dealing with 3D volumetric data common in modalities like CT and MRI scans. Recent benchmarks in this domain evaluate how well models can interpret complex spatial relationships, generate accurate clinical descriptions, and support diagnostic decision-making from 3D medical imagery. These evaluations reveal both the promise of extending MLLMs beyond 2D images and the significant challenges in handling the increased complexity of volumetric data. Benchmarks in this area assess diverse capabilities including anatomical understanding, report generation, visual question answering, and diagnostic accuracy.

M3D~\cite{bai_m3d_2024} represents a significant advancement in evaluating 3D medical image analysis capabilities through MLLMs. This benchmark comprises a large-scale dataset (M3D-Data) containing 120,000 image-text pairs and 662,000 instruction-response pairs specifically tailored for various 3D medical tasks. The benchmark evaluates models across eight diverse tasks in five major categories: image-text retrieval, report generation, visual question answering (both open-ended and closed-ended formats), positioning (including referring expression comprehension and generation), and segmentation (both semantic and referring expression segmentation). The researchers also introduce M3D-LaMed, a versatile multimodal model for 3D medical image analysis that integrates a 3D Vision Transformer as the vision encoder, a 3D spatial pooling perceiver for reducing embedding dimensions, and the LLaMA-2-7B model as the base LLM. Comprehensive evaluation demonstrates that M3D-LaMed outperforms existing solutions across the benchmark tasks, establishing a new state-of-the-art for 3D medical image understanding, while also showing promising capabilities in answering out-of-distribution questions.

TriMedLM~\cite{10822809} introduces a trimodal approach to medical imaging analysis, combining conventional image data with additional modalities to enhance diagnostic capabilities. This benchmark evaluates models on their ability to integrate information across these modalities, recognizing that medical diagnosis often requires synthesis of diverse data types. The model architecture employs a specialized design to handle the trimodal inputs, enabling more comprehensive analysis than is possible with unimodal approaches. Evaluation results demonstrate that this integrated approach yields improvements in diagnostic accuracy and completeness of analysis, highlighting the value of multimodal integration in medical imaging applications.

RadGPT~\cite{bassi2025radgptconstructing3dimagetext} presents an anatomy-aware vision-language AI agent for generating detailed reports from CT scans, with a particular focus on tumor analysis. The system first segments tumors (including benign cysts and malignant tumors) and their surrounding anatomical structures, then transforms this information into both structured and narrative reports containing detailed information on tumor size, shape, location, attenuation, volume, and interactions with surrounding blood vessels and organs. Evaluation on unseen hospitals demonstrated strong performance in tumor detection, with high sensitivity/specificity for small tumors (<2 cm): 80/73\% for liver tumors, 92/78\% for kidney tumors, and 77/77\% for pancreatic tumors. For larger tumors, sensitivity ranged from 89\% to 97\%, significantly surpassing the state-of-the-art in abdominal CT report generation. The researchers also created AbdomenAtlas 3.0, the first publicly available image-text 3D medical dataset, comprising over 1.8 million text tokens and 2.7 million images from 9,262 CT scans, including 2,947 tumor scans/reports of 8,562 tumor instances. This resource provides valuable capabilities including localization of tumors in organ sub-segments, determination of pancreatic tumor stage, and individual analyses of multiple tumors—a feature rarely found in human-made reports.

\subsubsection{Pathology/Microscopy}
Pathology and microscopy represent specialized areas of medical imaging that involve the analysis of cellular and tissue-level structures at various magnifications. Multimodal large language models applied to these domains must interpret fine-grained visual features that are substantially different from natural images or radiological scans, requiring both deep domain knowledge and sophisticated visual reasoning. Recent benchmarks in this area evaluate how effectively MLLMs can perform tasks ranging from basic microscopic image interpretation to complex diagnostic reasoning. These evaluations consistently reveal a significant performance gap between current state-of-the-art models and human expert pathologists, highlighting the challenges of adapting general-purpose vision-language models to this specialized domain.

Micro-Bench~\cite{lozano2024micro} introduces a benchmark containing 3,200 microscopy images paired with visual question answering tasks to evaluate models' understanding of microscopic structures and features. In evaluations, GPT-4V achieved 54.7\% accuracy, demonstrating moderate capabilities but still falling significantly short of expert-level performance. The benchmark specifically tests models' ability to interpret fine-grained visual features unique to microscopy images, including cellular morphology, tissue architecture, and subcellular structures that require specialized domain knowledge to properly analyze. This benchmark addresses the critical need for evaluating models on their ability to interpret scientific imagery that differs substantially from natural images seen in general visual reasoning tasks.

PathMMU~\cite{sun2024pathmmu} represents the largest expert-validated pathology benchmark for evaluating multimodal models, comprising 33,428 multiple-choice questions and 24,067 pathology images from diverse sources. Each question is accompanied by an explanation for the correct answer. The benchmark was constructed using a three-step process: data collection from diverse sources (PubMed, educational videos, pathology textbooks, expert Twitter posts), detailed description generation using GPT-4V to enhance image captions, and question generation with rigorous expert validation by seven pathologists who manually reviewed approximately 12,000 questions in the validation and test sets. Evaluation of 18 state-of-the-art models revealed that even the most advanced model, GPT-4V, achieved only 49.8\% accuracy, compared to 71.8\% achieved by human pathologists—a gap of over 20 percentage points. The benchmark also revealed concerning trends: 15 out of 18 tested models achieved no more than 40\% accuracy, models demonstrated unexpected robustness to image corruption (raising questions about whether they actually utilized visual information), and text-only models performed surprisingly well, suggesting models may take shortcuts rather than performing proper image analysis.

MicroVQA~\cite{burgess2025microvqamultimodalreasoningbenchmark} provides 2,800 question-answer pairs focused specifically on scientific microscopy, evaluating models' capability to reason about microscopic structures and processes across biological, medical, and materials science applications. GPT-4V achieved a modest 48.2\% accuracy on this benchmark, highlighting the difficulty of interpreting specialized scientific imagery. The benchmark is designed to test not only recognition of microscopic structures but also reasoning about their function, relationships, and significance in scientific contexts. This evaluation provides insights into the current limitations of multimodal models when applied to specialized scientific domains that require both visual understanding and domain-specific knowledge.

SlideChat~\cite{chen2025slidechatlargevisionlanguageassistant} introduces a dialogue-based approach to pathology slide analysis, enabling interactive exploration of pathological specimens. The system supports conversational interaction where users can ask questions about specific regions or features of a slide, approximating the interactive workflow of pathologists who iteratively examine different areas of a specimen at varying magnifications. This benchmark evaluates models on their ability to maintain context through a dialogue, accurately describe pathological features, and respond appropriately to follow-up questions—capabilities that more closely match how pathologists actually work with digital slides than single-turn question answering. The specialized SlideChat model demonstrates the value of architecture and training approaches tailored specifically for pathology applications.

$\mu$-Bench~\cite{lozano2024mubenchvisionlanguagebenchmarkmicroscopy} offers a comprehensive evaluation framework for microscopy image analysis across multiple tasks, including classification, segmentation, and visual question answering. The benchmark includes diverse microscopy modalities and sample types, providing a holistic assessment of models' capabilities in this domain. Evaluation results vary significantly across tasks and model architectures, with no single approach excelling across all microscopy analysis scenarios. This multi-task benchmark highlights the complexity of microscopy analysis and the need for models that can flexibly adapt to different analytical requirements, from basic feature identification to complex interpretative reasoning.

Medical imaging benchmarks consistently reveal the substantial gap between current model performance and expert-level capabilities, with even top-performing models like GPT-4V achieving accuracy well below human pathologists. A notable trend across these evaluations is models' tendency to rely more on textual information than properly analyzing visual content, taking shortcuts rather than performing true visual reasoning. Despite these limitations, specialized architectures like M3D-LaMed and RadGPT demonstrate the value of domain-specific training and architectural approaches, particularly for 3D data analysis. Future development should focus on improving fine-grained visual feature extraction, enhancing multi-scale analysis capabilities, and developing more rigorous evaluation protocols that prevent models from exploiting textual shortcuts.

\section{Language Understanding}\label{sec:language}
Language understanding benchmarks evaluate models' capabilities across diverse linguistic tasks, from multimodal integration to complex reasoning. These benchmarks reveal both significant progress and persistent challenges in developing truly comprehensive language understanding systems. An overview of all the relevant benchmarks is presented in Table~\ref{tbl:language_discipline}.

\begin{table*}[!ht]
  \caption{Language Understanding Discipline: Domain-Specific Benchmarks}\label{tbl:language_discipline}
  \centering
  
  \resizebox{\fulllength}{!}{%
  \begin{tabular}{lllcccccc} 
  \cmidrule[\heavyrulewidth]{1-9}
  \textbf{Domain} & \textbf{Sub-domain} & \textbf{Benchmark} & \textbf{Scale} & \textbf{Task Type} & \textbf{Input Modality} & \textbf{Model} & \textbf{Performance} & \textbf{Key Focus} \\
  \cmidrule[\heavyrulewidth]{1-9}
  \multirow{3}{*}{Multimodal Communication} 
  & \multirow{3}{*}{Visual-Language Integration}
  & LongLLaVA~\cite{wang_longllava_2024} & \textcolor{quantified}{1000 images} & Long-context & Multi-modal & LongLLaVA & \textcolor{unspecified}{N/A} & Hybrid architecture \\
  && LLaVA-OneVision~\cite{li2024llava} & \textcolor{unspecified}{N/A} & Multi-scenario & Multi-modal & LLaVA & \textcolor{unspecified}{N/A} & Cross-modal transfer \\
  && KOSMOS-1~\cite{huang2023languageneedaligningperception} & \textcolor{collection}{Web-scale} & Multi-task & Multi-modal & KOSMOS-1 & \textcolor{unspecified}{N/A} & Cross-modal transfer \\
  && KOSMOS-2~\cite{peng2023kosmos2groundingmultimodallarge} & \textcolor{collection}{Large-scale} & Multi-task & Multi-modal & KOSMOS-2 & \textcolor{unspecified}{N/A} & Visual grounding \\
  \cmidrule[\heavyrulewidth]{1-9}
  \multirow{1}{*}{Cross-lingual Tasks} 
  & \multirow{1}{*}{Multilingual Models}
  & ChatGLM~\cite{glm2024chatglm} & \textcolor{quantified}{10T tokens} & Multi-task & Multi-modal & GLM-4 & Matches GPT-4 & Chinese-English LLM \\
  && CVLUE~\cite{wang2025cvlue} & \textcolor{quantified}{30K+ samples} & Multi-task & Multi-modal & X\textsuperscript{2}VLM & 36-55\% & Chinese VL understanding \\
  \cmidrule[\heavyrulewidth]{1-9}
  \multirow{1}{*}{Reasoning} 
  & \multirow{1}{*}{Problem Solving}
  & ToT~\cite{yao2024tree} & \textcolor{unspecified}{N/A} & Problem solving & Text & GPT-4 & 74\% & Deliberate reasoning \\
  \cmidrule{2-9}
  & \multirow{1}{*}{Domain-Specific Applications}
  & AgEval~\cite{arshad2025leveraging} & \textcolor{quantified}{12 tasks} & Plant phenotyping & Multi-modal & Multiple & 46-73\% F1 & Agricultural tasks \\
  \cmidrule[\heavyrulewidth]{1-9}
  \end{tabular}%
  }
  
  \end{table*}

\subsection{Multimodal Communication}
Recent developments in visual-language integration have focused on addressing key challenges in multimodal large language models (MLLMs), particularly in handling long-context scenarios and improving cross-modal transfer capabilities. The benchmarks in this domain demonstrate significant progress in expanding the contextual understanding and efficiency of MLLMs.

A notable advancement in long-context processing is demonstrated by LongLLaVA~\cite{wang_longllava_2024}, which introduces a hybrid architecture combining Mamba and Transformer blocks. This approach enables processing of up to 1000 images on a single GPU while maintaining competitive performance. The architecture addresses the dual challenges of performance degradation with increased image count and computational efficiency, representing a step forward in scaling MLLMs for practical applications.

Cross-modal transfer has emerged as another critical focus area, as evidenced by LLaVA-OneVision~\cite{li2024llava} and KOSMOS-1~\cite{huang2023languageneedaligningperception}. LLaVA-OneVision demonstrates the feasibility of a unified model capable of handling multiple visual scenarios, including single-image, multi-image, and video understanding. This approach enables strong transfer learning across different modalities, particularly from images to videos. Similarly, KOSMOS-1 showcases the potential of web-scale training for developing MLLMs that can perceive general modalities and perform few-shot learning and instruction following.

As shown in Table below, these benchmarks vary in their scale and focus. While LongLLaVA provides a quantified dataset of 1000 images, KOSMOS-1 operates at a web-scale level. The progression from single-task to multi-task and multi-scenario capabilities indicates the field's movement toward more versatile and robust models. However, challenges remain in balancing computational efficiency with model performance and ensuring effective knowledge transfer across different modalities.
\subsection{Cross-lingual Tasks}

In the domain of multilingual models, ChatGLM~\cite{glm2024chatglm} represents a significant advancement in cross-lingual capabilities. The model family, particularly GLM-4, demonstrates the potential of large-scale multilingual training with a corpus of 10 trillion tokens primarily in Chinese and English, along with coverage of 24 additional languages. As shown in Table~\ref{tbl:language_discipline}, ChatGLM achieves performance comparable to GPT-4 across various benchmarks while excelling specifically in Chinese language tasks. The model's success in matching GPT-4's performance while maintaining strong multilingual capabilities suggests the effectiveness of their multi-stage post-training process, which combines supervised fine-tuning with learning from human feedback.

CVLUE~\cite{wang2025cvlue} addresses a critical gap in vision-language understanding evaluation for Chinese culture. Unlike existing Chinese VL datasets that often use Western-centric images from English datasets, CVLUE features images specifically selected by Chinese native speakers to represent Chinese cultural contexts. The benchmark comprises four distinct tasks: image-text retrieval, visual question answering, visual grounding, and visual dialogue. Evaluations of multilingual VLMs on CVLUE reveal significant performance gaps between English and Chinese vision-language understanding capabilities, with models scoring 20-35\% lower on Chinese tasks compared to English counterparts. The benchmark's category-level analysis demonstrates that fine-tuning on Chinese culture-related visual-language data substantially improves model performance, particularly for culturally specific concepts.

\subsection{Reasoning}

The field of reasoning has seen methodological innovations in how language models approach complex problem-solving tasks. A notable advancement in this domain is the Tree of Thoughts (ToT) framework~\cite{yao2024tree}, which extends beyond traditional sequential reasoning approaches. As indicated in Table~\ref{tbl:language_discipline}, ToT demonstrates substantial improvements in deliberate problem solving, achieving a 74\% success rate on specific tasks where conventional methods struggle. The framework enables language models to explore multiple reasoning paths simultaneously, implement strategic lookahead, and perform self-evaluation of intermediate decisions. This approach particularly excels in tasks requiring non-trivial planning or search capabilities, such as mathematical puzzles and creative writing tasks, where initial decisions significantly impact the final outcome. The success of ToT suggests the importance of structured exploration and deliberate decision-making in enhancing language models' problem-solving capabilities.

\subsection{Domain-Specific Applications}
AgEval~\cite{arshad2025leveraging} introduces a comprehensive benchmark for evaluating Vision Language Models (VLMs) on specialized agricultural tasks, particularly plant stress phenotyping. The benchmark encompasses 12 diverse plant stress tasks across identification, classification, and quantification categories, assessing zero-shot and few-shot in-context learning performance of models like Claude, GPT, Gemini, and LLaVA. Results demonstrate VLMs' rapid adaptability to specialized tasks, with F1 scores increasing from 46.24\% to 73.37\% in 8-shot settings. The study also quantifies performance disparities across plant stress classes using coefficient of variation metrics (ranging from 26.02\% to 58.03\%) and shows that strategic example selection improves model reliability by up to 15.38\%. AgEval establishes that VLMs, with minimal few-shot examples, can serve as viable alternatives to traditional specialized models in agricultural applications.

\section{Conclusion and Future Work}
This survey has underscored the indispensable role of domain-specific benchmarks in guiding Multimodal Large Language Models (MLLMs) beyond general proficiency to achieve the critical `last mile problem' of efficacy in specialized applications. While foundational models show broad capabilities, their performance often diminishes when faced with the nuanced demands of fields like finance, where even top models like GPT-4 struggle on benchmarks such as FinanceBench~\cite{islam2023financebench}, or in medicine, where models can be confused by visually distinct images as seen in MediConfusion~\cite{sepehri_mediconfusion_2024}. The path forward necessitates a concerted effort towards greater standardization and comparability in benchmark design and evaluation protocols. Future benchmarks must evolve to incorporate comprehensive metrics beyond mere accuracy, assessing robustness, efficiency, and safety. They also need to become more dynamic, or ``living,'' to keep pace with rapid MLLM advancements, and must challenge models with increasing multimodal complexity, noisy real-world data, and tasks that demand deeper, more causal reasoning than currently evaluated—moving beyond the current limitations where, for instance, even advanced models achieve only moderate success on comprehensive medical benchmarks like GMAI-MMBench~\cite{ye2024gmai}.

Beyond refining evaluation quality, future work must focus on enhancing the broader applicability and trustworthiness of MLLMs. This includes a critical emphasis on ethical considerations and societal impact, developing benchmarks that rigorously assess for biases, privacy vulnerabilities, and the potential for misuse, particularly in sensitive areas like social media analysis as highlighted by concerns in MM-Soc~\cite{jin2024mm} and Priv-IQ~\cite{shahriar_priv-iq_2025}, or within autonomous systems. The `black box' nature of MLLMs remains a significant hurdle; thus, fostering interpretability and explainability through benchmark design is paramount, especially in high-stakes domains like healthcare where models sometimes resort to textual shortcuts rather than true visual reasoning, as observed in pathology benchmarks like PathMMU~\cite{sun2024pathmmu}. Furthermore, advancing MLLM utility involves exploring cross-domain knowledge transfer and integrating human-in-the-loop evaluations for tasks requiring subjective or nuanced judgment, prevalent in the humanities and arts where automated metrics are often insufficient. Addressing these multifaceted challenges will require a holistic approach to benchmark development, guiding MLLMs to be not only powerful but also reliable, interpretable, and ethically aligned with societal values as they transform a myriad of applications.

\bibliographystyle{elsarticle-num}
\bibliography{references, ref}


\end{document}